\documentclass[11pt]{article} 
\usepackage{graphicx}
\usepackage{xcolor}
\usepackage{algorithm}
\usepackage{algpseudocode}
\usepackage{lineno}
\usepackage[utf8]{inputenc}
\usepackage{authblk}
\usepackage[normalem]{ulem}
\usepackage{fullpage, times}
\usepackage[parfill]{parskip}
\usepackage{color,colortbl}
\definecolor{mydarkblue}{rgb}{0,0.08,0.45}
\usepackage[colorlinks=true,
    linkcolor=mydarkblue,
    citecolor=mydarkblue,
    filecolor=mydarkblue,
    urlcolor=mydarkblue,
    pdfview=FitH]{hyperref}
\usepackage{mdframed}
\usepackage{tikz}
\usepackage{caption}
\usetikzlibrary{decorations.pathreplacing} 

\usepackage{microtype}
\usepackage{graphicx}
\usepackage{natbib}
\usepackage{authblk}

\usepackage{booktabs} 
\usepackage{subcaption}
\usepackage{multirow}
\usepackage{makecell}
\usepackage[role = cvprblue, grid = lightgrey]{credits}
\usepackage{xcolor}
\definecolor{cvprblue}{rgb}{0.21,0.49,0.74}
\definecolor{lightgrey}{RGB}{150,150,150}

\usepackage{hyperref}

\usepackage{amsmath}
\usepackage{amssymb}
\usepackage{mathtools}
\usepackage{amsthm}
\usepackage[subtle]{savetrees}
\usepackage{float}
\usepackage{placeins}

\usepackage[capitalize,noabbrev]{cleveref}

\theoremstyle{plain}
\newtheorem{theorem}{Theorem}[section]

\theoremstyle{definition}

\theoremstyle{remark}

\usepackage[textsize=tiny]{todonotes}

\newcommand{\R}{\mathbb{R}}
\DeclareMathOperator*{\argmin}{arg\,min}

\newcommand\sref{\S\ref}
\newcommand\eref{Eq.~\ref}
\newcommand\fref{Fig.~\ref}
\newcommand\tref{Tab.~\ref}

\newcommand{\update}[1]{{{#1}}}
\newmdtheoremenv{mddef}{Definition}

\usepackage{url}
\usepackage[labelfont=bf]{caption}
\usepackage{soul}

\title{Understanding and Mitigating Distribution Shifts \\ For Machine Learning Force Fields}

\makeatletter
\renewcommand{\@fnsymbol}[1]{\ensuremath{\ifcase#1\or a\or b\or c\or d\else\@ctrerr\fi}}
\makeatother

\author[1]{Tobias Kreiman\thanks{Corresponding author: \texttt{tkreiman@berkeley.edu}}}
\author[1, 2]{Aditi S. Krishnapriyan}

\affil[1]{UC Berkeley}
\affil[2]{LBNL}

\date{}

\newcommand{\remove}[1]{}

\begin{document}

\maketitle

\begin{abstract}
Machine Learning Force Fields (MLFFs) are a promising alternative to expensive \textit{ab initio} quantum mechanical molecular simulations. Given the diversity of chemical spaces that are of interest and the cost of generating new data, it is important to understand how MLFFs generalize beyond their training distributions. \update{In order to characterize and better understand distribution shifts in MLFFs}, we conduct diagnostic experiments on chemical datasets, revealing common shifts that pose significant challenges, even for large foundation models trained on extensive data. Based on these observations, we hypothesize that current supervised training methods inadequately regularize MLFFs, resulting in overfitting and learning poor representations of out-of-distribution systems. We then propose two new methods as initial steps for mitigating distribution shifts for MLFFs. Our methods focus on test-time refinement strategies that incur minimal computational cost \update{and do not use expensive \textit{ab initio} reference labels}. The first strategy, based on spectral graph theory, modifies the edges of test graphs to align with graph structures seen during training. 
Our second strategy improves representations for out-of-distribution systems at test-time by taking gradient steps using an auxiliary objective, such as a cheap physical prior. Our test-time refinement strategies significantly reduce errors on out-of-distribution systems, suggesting that MLFFs are capable of and can move towards modeling diverse chemical spaces, but are not being effectively trained to do so. Our experiments establish clear benchmarks for evaluating the generalization capabilities of the next generation of MLFFs. Our code is available at \url{https://tkreiman.github.io/projects/mlff_distribution_shifts/}.
\end{abstract}

\vspace{-8pt}
\section{Introduction}
\label{sec:intro}
\vspace{-5pt}

Understanding the quantum mechanical properties of atomistic systems is crucial for the discovery and development of new molecules and materials. Computational methods like Density Functional Theory (DFT) are essential for studying these systems, but the high computational demands of such methods limit their scalability. Machine Learning Force Fields (MLFFs) have emerged as a promising alternative, learning to predict energies and forces from reference quantum mechanical calculations. MLFFs are faster than traditional \textit{ab initio} methods, and their accuracy is rapidly improving for modeling complex atomistic systems~\citep{Batzner_2022nequip, schutt_schnet_2017, gasteiger_gemnet_2021, batatia_mace_2022}.

Given the computational expense of \textit{ab initio} simulations for all chemical spaces of interest, there has been a push to train larger and more accurate MLFFs, designed to work well across many different systems. Developing models with general representations that accurately capture diverse chemistries has the potential to reduce or even eliminate the need to recollect data and retrain a model for each new system. To determine which systems an MLFF can accurately describe and to assess the reliability of its predictions, it is important to understand how MLFFs generalize beyond their training distributions. This understanding is essential for applying MLFFs to new and diverse chemical spaces, ensuring that they perform well not only on the data they were trained on, but also on unseen, potentially more complex systems.

We conduct an in-depth exploration to identify and understand distribution shifts. On example chemical datasets, we find that many large-scale models struggle with common distribution shifts \citep{kovács2023maceoff23, shoghi2023jmp, liao2024equiformerv2, batatia2024macemp} (see \sref{sec:distribution_shifts}). These generalization challenges suggest that current supervised training methods for MLFFs overfit to training distributions and do not enable MLFFs to generalize accurately.  
We demonstrate that there are multiple reasons that this is the case, including challenges associated with poorly-connected graphs and learning unregularized representations, evidenced by jagged predicted potential energy surfaces for out-of-distribution systems.    

Building on our observations, we take initial steps to mitigate distribution shifts for MLFFs \update{without test set reference labels} by proposing two approaches: test-time radius refinement and test-time training \citep{sun2020testtime, gandelsman2022testtime, jang2023testtime}. For test-time radius refinement, we modify the construction of test-graphs to match the training Laplacian spectrum, overcoming differences between training and testing graph structures. 
For test-time training (TTT), we address distribution shifts by taking gradient steps on an auxiliary objective at test time. \update{Analogous to self-supervised objectives in computer vision TTT works \citep{gandelsman2022testtime, sun2020testtime, hardt2024test}, we use an efficient prior as a target to improve representations at test time.}

Although completely closing the out-of-distribution to in-distribution gap remains a challenging open machine learning problem \citep{sun2020testtime, gandelsman2022testtime}, our extensive experiments show that our test-time refinement strategies are effective in mitigating distribution shifts for MLFFs. Our experiments demonstrate that low quality data can be used to improve generalization for MLFFs, and they establish clear benchmarks that highlight ambitious but important generalization goals for the next generation of MLFFs.

\update{We summarize our main contributions here:}
\update{
\vspace{-8pt}
\begin{enumerate}
\itemsep-1pt 
    \item We run diagnostic experiments on different chemical datasets to characterize and understand common distribution shifts for MLFFs in \sref{sec:distribution_shifts}.
    \item Based on (1), we take first steps at mitigating MLFF distribution shifts in \sref{sec:methods} with two test-time refinement strategies. 
    \item The success of these methods, validated through extensive experiments in \sref{sec:experiments}, suggests that MLFFs are not being adequately trained to generalize, despite current models having the expressivity to close the gap on the distribution shifts explored in \sref{sec:distribution_shifts}.  
\end{enumerate}}

\section{Related Work}
\label{sec:related}

\paragraph{Distribution Shifts.} There is a long line of literature studying distribution shifts in the machine learning community, which we briefly summarize here. \citet{sugiyama07_domainshift} demonstrated how to perform importance weighted cross validation to perform model selection under distribution shifts. Methods have been proposed to measure and improve the robustness of models to distribution shifts in images \citep{taori2020_imgds, zhao2022oodimg} and language \citep{ds_language}. Numerous methods have been proposed to tackle distribution shifts including, but not limited to, techniques based on meta learning \citep{NEURIPS2020_oodmaml} and ensembles \citep{Zhou_2021_dsensemble}.

Recent work has also begun identifying generalization challenges with MLFFs \citep{Li2025_oodmat, Bihani2024_egraff}. \citet{deng2024overcomingsystematicsofteninguniversal} find that MLFFs systematically underpredict energy surfaces, and that this underprediction can be ameliorated with a small number of fine-tuning steps on reference calculations. Our experiments complement these initial findings of underestimation, and we also identify other types of distribution shifts, like connectivity and atomic feature shifts. Our proposed test-time refinement solutions are also able to mitigate distribution shifts \textit{without} any reference data, and they provide insights into \textit{why} MLFFs are unable to generalize.

\paragraph{Multi-Fidelity Machine Learning Force Fields.}

\citet{Behler2007} popularized the use of machine learning for modeling force fields, leading to numerous downstream applications \citep{Artrith2011} and refinements to model increasingly complicated systems \citep{Drautz2019}. More recent work has explored training MLFFs with observables \citep{Fuchs2025_obs, raja2025stabilityaware, Han2025}, distilling MLFFs with physical constraints \citep{amin2025towards}, and using multiple levels of theory during training. \citet{amin2025towards} found that knowledge distillation can enable smaller models to outperform larger models in certain specialized tasks, suggesting that the larger MLFFs may not have been trained in a way that fully leverages their capacity. \citet{Jha2019_transfer}, \citet{Gardner2024_syntheticcarbon}, and \citet{shui2022injecting_domainknowledge} leveraged cheap or synthetic data to improve data efficiency and accuracy. \citet{deltalearning} popularized the $\Delta$-learning approach \citep{Bogojeski2020_delta}, where a model learns to predict the difference between some prior and the reference quantum mechanical targets. Multi-fidelity learning generalizes $\Delta$-learning by building a hierarchy of models that predict increasingly accurate levels of theory \citep{multifidelity_2023,Vinod_2023multifidelity2, Forrester2007_multifidelity, Heinen2024-multifidelity}. Making predictions in the hierarchical multi-fidelity setting corresponds to evaluating a baseline fidelity level and then refining this prediction with models that provide corrections to more accurate levels of theory in the hierarchy. 

Our work differs from these works in several ways. We focus on developing training strategies that address distribution shifts. In contrast to prior multi-fidelity works, we learn \textit{representations} from multiple levels of theory using pre-training, fine-tuning, and joint-training objectives. Rather than fine-tuning all the model weights like in \citet{Jha2019_transfer}, \citet{Gardner2024_syntheticcarbon}, and  \citet{shui2022injecting_domainknowledge}, we explore freezing and regularization techniques that enable test-time training. Our new test-time objectives update the model's representations when faced with out-of-distribution examples, improving performance on out-of-distribution systems. Multi-fidelity approaches by themselves do not tackle the challenge of transferring to new, unseen systems at test-time. Nevertheless, combining our training strategies with other multi-fidelity approaches presents an interesting direction for future work.     

\paragraph{Test-Time Training.}
The test-time training (TTT) framework adapts predictive models to new test distributions by updating the model at test-time with a self-supervised objective \citet{sun2020testtime}. \citet{sun2020testtime} demonstrated that forcing a model to use features learnt from a self-supervised objective during the main task allows the model to adapt to out-of-distribution examples by tuning the self-supervised objective. Follow up work showed the benefits of TTT across computer vision and natural language processing, exploring a range of self-supervised objectives \citep{gandelsman2022testtime, jang2023testtime, hardt2024test}.

\vspace{-5pt}
\section{Distribution Shifts for Machine Learning Force Fields} 
\label{sec:distribution_shifts}

\begin{figure*}
    \centering
    \includegraphics[width=0.75\linewidth]{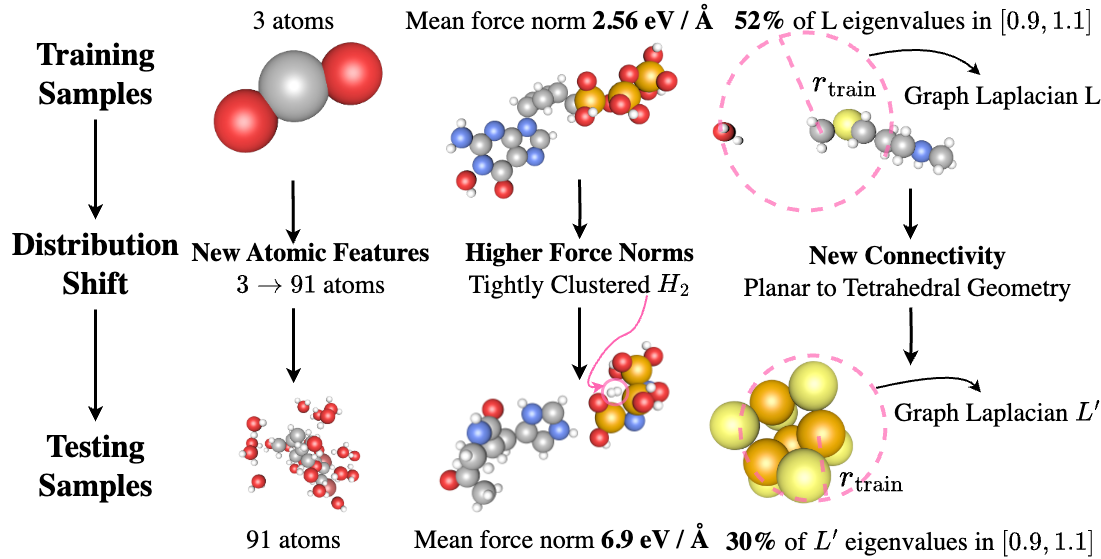}
    \caption{\textbf{Distribution Shifts for MLFFs.} We visualize distribution shifts based on changes in features, labels, and graph structure.  Typical training samples from SPICE~\cite{Eastman2023spice} and new systems from SPICEv2 \citep{eastman2024spice2} are displayed. An atomic feature shift is illustrated by comparing a three-atom molecule with a larger molecular system containing 91 atoms (left). A force norm shift is shown by the close proximity of an $H_2$ molecule (circled in pink), leading to high force norms \update{(middle)}. A connectivity shift is shown by the tetrahedral geometry in $P_4S_6$, which differs from the typical planar geometry seen during training (right).}  
    \label{fig:dist_shifts}
\vspace{-12pt}
\end{figure*}

\subsection{Problem Setup and Background}
\label{sec:problem_setup}

MLFFs approximate molecule-level energies and atom-wise forces for a chemical structure by learning neural network parameters from data. For a given a molecular structure, the input to the ML model consists of two vectors: $\mathbf{r} \in \R^{n \times 3}$, $\mathbf{z} \in \R^{n \times d}$, where $n$ represents the number of atoms in the molecule, $\mathbf{r}$ are the atomic positions, and $\mathbf{z}$ are the features of the atom, such as atomic numbers or whether an atom is fixed or not. The model outputs $\hat{E} \in \R, \mathbf{\hat{F}} \in \R^{n \times 3}$, which are the predicted total potential energy of the molecule and the predicted forces acting on each atom.
The learning objective is typically formulated as a supervised loss function, which  measures the discrepancy between the predicted energies and forces and reference energies and forces:
\begin{equation}
    \label{eqn:supervised_loss}
    \vspace{-6pt}
    \mathcal{L}(\mathbf{F}, E) =  \lambda_{E} ||E_{ref} - \hat{E}||_2^2 + \lambda_{F} \sum_{i=1}^n ||\mathbf{F_{i, ref}} - \mathbf{\hat{F}_i}||_2^2,
\vspace{-6pt}
\end{equation}
where $\lambda_E, \lambda_F$ are hyperparameters.

Most modern MLFFs are implemented as graph neural networks (GNNs)~\cite{gilmer2017neural}. 
Consequently, $\hat{E}$ and $\mathbf{\hat{F}}$ are functions of $\mathbf{z}$, $\mathbf{r}$, and $A \in \R^{n \times n}$, the adjacency matrix representing the molecule: 
\begin{equation}
    \label{eqn: gnn}
    \hat{E}, \mathbf{\hat{F}} = f(\mathbf{z}, \mathbf{r}, A)
\end{equation}
The atoms in the molecule are modeled as nodes in a graph, and edges are specified by the adjacency matrix that includes connections to all atoms within a specified radius cutoff~\citep{gasteiger_gemnet_2021, batatia_mace_2022}. The adjacency matrix fully determines a graph structure, and thus defines the graph over which the GNN performs its computation.

\subsection{Criteria for Identifying Distribution Shifts}
\label{sec:criteria_ds}

In this section, we formalize criteria for identifying distribution shifts based on the features, labels, and graph structures in chemical datasets. We define these distribution shifts broadly to encompass the diversity of chemical spaces. We also note that distribution shifts can occur independently along each dimension: e.g., a shift in features does not necessarily imply a shift in labels (see \sref{apx:distribution_shifts} for details). This categorization provides a framework for understanding the types of distribution shifts an MLFF may encounter (see \fref{fig:dist_shifts}). \update{This understanding motivates the refinement strategies described in \sref{sec:methods} that take first steps at mitigating these shifts, providing insights into why MLFFs are susceptible to these shifts in the first place.}

\begin{figure}
    
\centering
\includegraphics[width=0.9\linewidth]{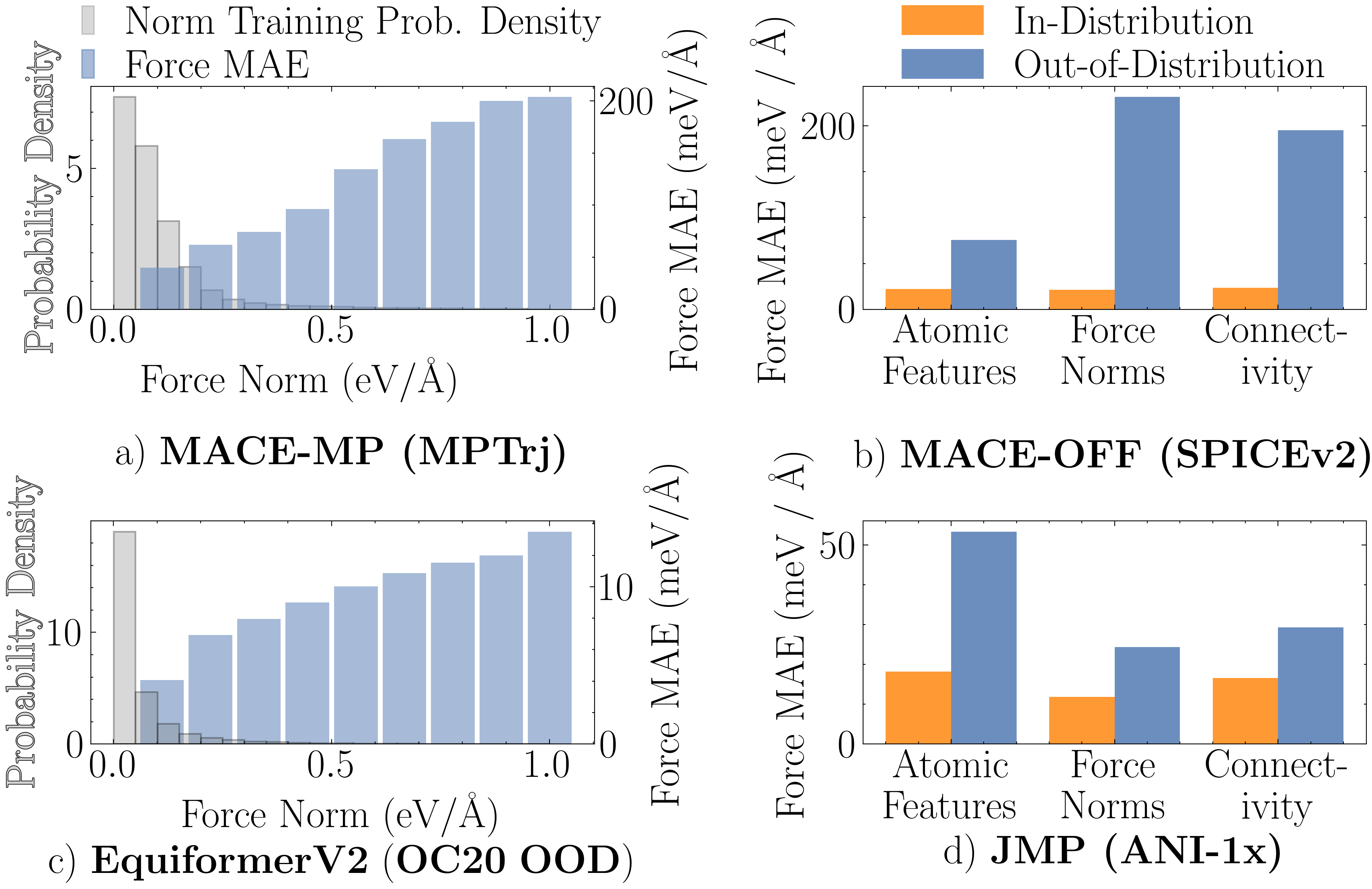}
\captionof{figure}{\textbf{Distribution Shifts for Large Models.} We study distribution shifts on four of the largest open-source MLFFs designed for broad chemical spaces. (a) We evaluate MACE-MP on the MPTrj train set. (b) We evaluate MACE-OFF on 10k new molecules from SPICEv2. \update{(c) We evaluate EquiformerV2 on the OC20 out-of-distribution validation set. (d) We evaluate JMP on the ANI-1x test set.} A molecule is considered out-of-distribution if it is more than 1 standard deviation away from the mean training force norm, system size, or connectivity (with respect to the spectral distance defined above \sref{sec:criteria_ds}). Despite their scale, these large foundation models have $2-10\times$ larger force mean absolute errors (MAE) when encountering distribution shifts.}
\label{fig:foundation_bad}
\end{figure}
 
\paragraph{Distribution Shifts in Atomic Features ($\mathbf{z}$).} 
Distribution shifts in atomic features refer to any change in the atomic composition of a chemical system. This includes, but is not limited to, cases where models are trained on systems containing mixtures of organic elements but tested on structures composed solely of carbon, or cases where there is a shift in system size between training and testing. \fref{fig:dist_shifts} illustrates an atomic feature distribution shift by comparing a small carbon dioxide molecule to larger molecular system containing 91 atoms. 


\vspace{-8pt}
\paragraph{Distribution Shifts in Forces ($\mathbf{F}$).}
An MLFF may also encounter a distribution shift in the force labels it predicts. A model trained on structures close to equilibrium, with low force magnitudes, might be tested on a structure with higher force norms. \fref{fig:dist_shifts} shows an example of a tightly clustered $H_2$ molecule, which leads to a force norm distribution shift.

\vspace{-8pt}
\paragraph{Distribution Shifts in Graph Structure and Connectivity ($A$).} Since many MLFFs are implemented as GNNs, they may encounter distribution shifts in the graph structure defined by $A$. We refer to these as connectivity distribution shifts because $A$ determines the graph connectivity used by the GNN. Connectivity distribution shifts are particularly common in molecular datasets, where one could encounter a benzene ring at test time, despite only having trained on long acyclic structures. \fref{fig:dist_shifts} provides an example of a connectivity distribution shift, going from planar training structures to a tetrahedral geometry at test time. 

We identify connectivity distribution shifts by analyzing the eigenvalue spectra of the normalized graph Laplacian:
\begin{equation}
    \label{eqn: graph_laplacian}
    L = I - (D)^{-\frac{1}{2}}A(D)^{-\frac{1}{2}},
\end{equation}
where $D \in \R^{n \times n}$ is the degree matrix ($D_{ii} = \text{degree}(\text{node}_i)$ and $D_{ij} = 0$ for $i \neq j$, $A_{ij} = 1$ if $||r_i - r_j||_2 \leq r_{cutoff}$ and $0$ otherwise), and $I$ is the identity. $L$ has eigenvalues $\lambda_0, \leq \lambda_1, \leq \dots \leq \lambda_{n-1}$, where $\lambda_i \in [0, 2] \; \forall i$, and the multiplicity of the $0$ eigenvalue equals the number of connected components in the graph.

Following previous work \citep{Chung1996_spectral_graph_theory, WILSON_spectral_norm}, we can compare structural differences between graphs by using the spectral distance \citep{Jovanovi2012_spectral_dist}. Since Laplacian spectra are theoretically linked to information propagation in GNNs \citep{WILSON_spectral_norm, digiovanni2023oversquashingmessagepassingneural}, the spectral distance is a natural choice for comparing molecular graphs (see \sref{sec: ttrr} and \sref{apx:theory_rr} for more details).     

\vspace{-8pt}
\paragraph{Observed Distribution Shifts for Large Models.} We contextualize the aforementioned distribution shifts by considering four large models: MACE-OFF, MACE-MP, EquiformerV2, and JMP \citep{kovács2023maceoff23, shoghi2023jmp, liao2024equiformerv2, batatia2024macemp} MACE-OFF is a 4.7M biomolecules foundation model trained on 951k structures primarily from the SPICE dataset \citep{Eastman2023spice}. The 15M parameter MACE-MP foundation model is trained on 1.5M structures from the Materials Project \citep{dengmptrj}. EquiformerV2 is a 150M parameter model trained on 100M+ structures from OC20 \citep{Chanussot_2021_oc20}. The JMP model has 240M parameters and is trained on 100M+ structures from OC20, OC22, ANI-1x, and Transition-1x \citep{Chanussot_2021_oc20, Tran_2023_oc22, Smith2020_ani1x, schreiner2022transition1x}. These models represent four of the largest open-source MLFFs to date, and they have been trained on some of the most extensive datasets available. We focus on these models since their scale is designed for tackling broad chemical spaces.

We examine the generalization ability of MACE-OFF by testing it on 10k new molecules from the SPICEv2 dataset \citep{eastman2024spice2} not included in the MACE-OFF training set. A molecule is defined as out-of-distribution if it is more than 1 standard deviation away from the mean training data force norm, system size, or connectivity (with respect to the spectral distance defined above \sref{sec:criteria_ds}). Despite its scale, MACE-OFF performs worse by an order of magnitude on out-of-distribution systems (see \fref{fig:foundation_bad}a). \update{We also evaluate JMP on the ANI-1x \citep{Smith2020_ani1x} test set defined in \citet{shoghi2023jmp}. JMP also suffers predictably from force norm, connectivity, and atomic feature distribution shifts (see \fref{fig:foundation_bad}d).}

We focus on force norm distribution shifts for MACE-MP and EquiformerV2, since connectivity is more uniform across bulk materials and catalysts, where atoms are packed tightly into a periodic cell. For MACE-MP, we evaluate its performance directly on the entire MPTrj dataset. This model does not have a clear validation set, as it was trained on all of the data to maximize performance \citep{batatia2024macemp}. MACE-MP still clearly performs worse as force norms deviate from the majority of the training distribution (see \fref{fig:foundation_bad}b). The performance deterioration would be more severe with a held-out test set. \update{EquiformerV2 also struggles with high force norm structures when evaluated on the validation out-of-distribution set from OC20 \citep{Chanussot_2021_oc20} (see \fref{fig:foundation_bad}c).}             

\paragraph{Observations.}  
Training larger models with more data is one approach to address these distribution shifts (for example, with active learning \citep{Vandermause2020activelearning, Kulichenko2024}). However, doing so can be computationally expensive. Our diagnostic experiments also indicate that scale alone might not fully address distribution shifts, as naively adding more in-distribution data does not help large models generalize better (see \fref{fig:foundation_bad}). The diversity of chemical spaces makes it exceedingly difficult to know the exact systems that an MLFF will be tested on \textit{a priori}, making it challenging to curate the perfect training set. These observations lead us to develop strategies that mitigate distribution shifts by modifying the training and testing procedure of MLFFs. Importantly, these refinement strategies can be combined with any further architecture and data advances. 
\vspace{-5pt}
\section{Mitigating Distribution Shifts with Test-Time Refinement Strategies for Machine Learning Force Fields} 
\label{sec:methods}

Based on the generalization challenges for foundation models (see \sref{sec:distribution_shifts}), we hypothesize that many MLFFs are severely overfitting to the training data, resulting in a failure to learn generalizable representations. Building on our observations in~\sref{sec:distribution_shifts} and to test this hypothesis, we develop two test-time refinement strategies that also mitigate distribution shifts. \update{We focus on test time evaluations, i.e., with access to test molecular structures but without access to reference labels.} First, by studying the graph Laplacian spectrum, we investigate how MLFFs, and GNNs in general \citep{bechlerspeicher2024gnnregular}, tend to overfit to the regular and well-connected training graphs. In \sref{sec: ttrr}, we address connectivity distribution shifts by aligning the Laplacian eigenvalues of a test structure with the connectivities of the training distribution. Second, we show that MLFFs are inadequately regularized, resulting in poor representations of out-of-distribution systems. We incorporate inductive biases from a cheap physical prior using our pre-training and test-time training procedure (\sref{sec: ttt}) to regularize the model and learn more general representations, evidenced by smoother predicted potential energy surfaces. The effectiveness of these test-time refinement strategies, validated through extensive experiments in \sref{sec:experiments} and \sref{apx:more_ttt}, may indicate that MLFFs are currently poorly regularized and overfit to graph structures seen during training, hindering broader generalization.

\vspace{-10pt}
\subsection{Test-Time Radius Refinement}
\label{sec: ttrr}

\FloatBarrier
\noindent
\begin{minipage}{0.42\textwidth}
  We hypothesize that MLFFs tend to overfit to the specific graph structures encountered during training. We can characterize graph structures by studying the Laplacian spectrum of a graph. At test time, we can then identify when an MLFF encounters a graph with a Laplacian eigenvalue distribution that significantly differs from the training graphs (see \ref{sec:criteria_ds}). To address this shift, we propose updating the test graph to more closely resemble the training graphs, thereby mitigating connectivity distribution shifts. Since the adjacency matrix $A$ and graph Laplacian $L$ are typically generated by a radius graph, we refine the radius cutoff at test time. Instead of using a fixed radius cutoff $r_{\text{train}}$ for both training and testing, adjusting the radius cutoff at test time can help achieve a connectivity that more closely resembles the training graphs. 
\end{minipage}
\hspace{0.01\textwidth}
\begin{minipage}{0.57\textwidth}

\centering
  \includegraphics[width=\linewidth]{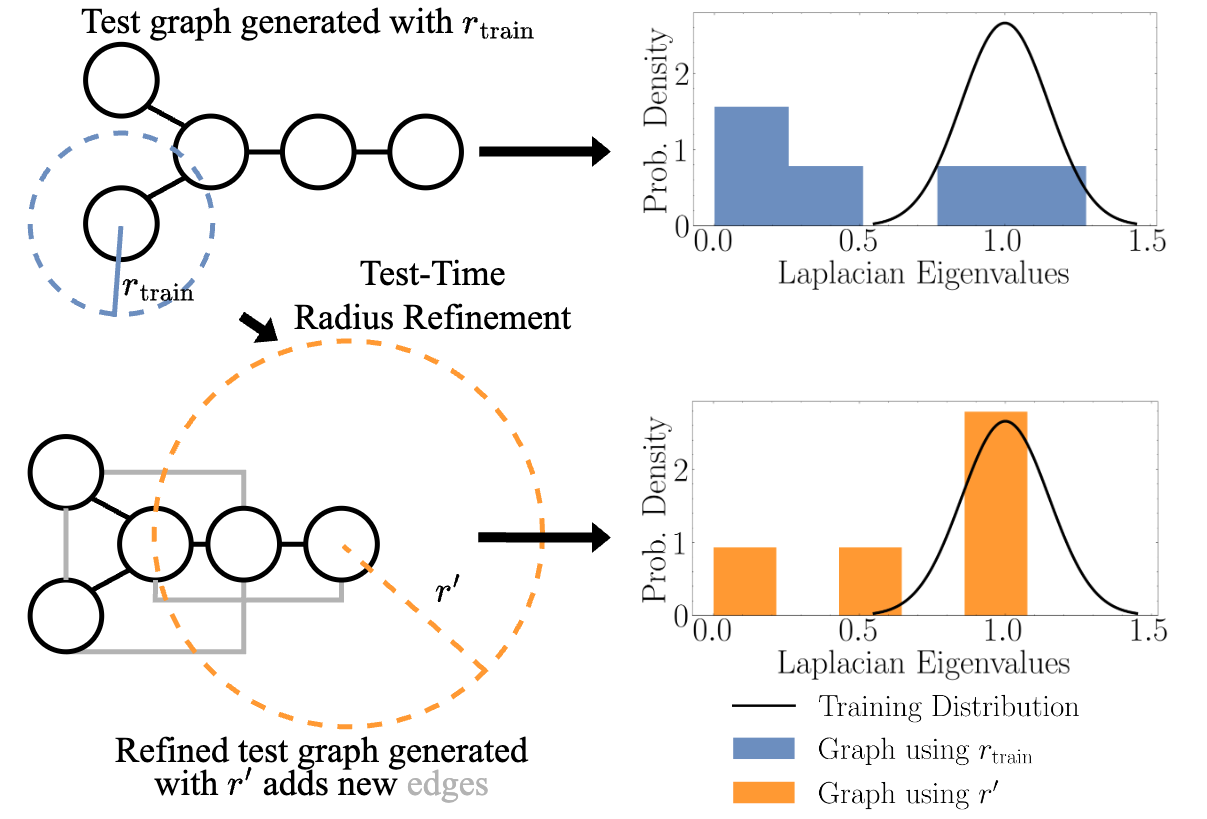}
  \captionof{figure}{\textbf{Test-Time Radius Refinement.} MLFFs tend to overfit to the well-connected graphs seen during training, which can be identified by the clustering of Laplacian eigenvalues around 1. To mitigate connectivity distribution shifts at test time, we find the optimal radius cutoff, which aligns the Laplacian eigenvalues of test graphs with those of the training distribution.}
  \label{fig:tt_rr}
 
\end{minipage}

Formally, for each test structure $j$, we search over $k$ new radius cutoffs $[r_i]_{i=1}^{k}$, calculate the new eigenvalue spectra for $L^{(j)}$ induced by the new cutoff $r_i$, and select the $r_i$ that minimizes the difference between the eigenvalue spectra of the new graph and the training graphs (see \fref{fig:tt_rr}): 
\begin{equation}
\label{eqn:rr}
    r^{(j)}_{\text{test}} = \argmin_{[r_i]_{i=1}^k} D(\lambda_{\text{train}}, \lambda(L^{(j)}(r_i))),
\end{equation}
where $\lambda_{\text{train}}$ is the training distribution of eigenvalues, $\lambda(L^{(j)}(r_i))$ is the Laplacian spectrum for sample $j$ generated with radius cutoff $r_i$, and $D$ is some distance function. We choose the squared spectral distance:
\begin{equation}
    \label{eqn:spectral_distance}
    D(\lambda_{\text{train}}, \lambda(L^{(j)}(r_i))) = \sum_k (\Bar{\lambda}_k - \lambda(L^{(j)}(r_i))_k)^2,
\end{equation}
where, following previous work, $\Bar{\lambda}$ is the average Laplacian spectrum of the training distribution with spectra padded with zeros to accomodate different sized graphs \citep{Chung1996_spectral_graph_theory, Jovanovi2012_spectral_dist}. While averaging the training distribution provides a lossy representation of the training connectivities, it is computationally impractical to compare each new test structure to all training graphs individually. One alternative is to count the number of training graphs within a certain cutoff of the spectral distance to assess how far a test graph is from the training distribution. However, this measure is highly correlated with the simpler spectral distance metric,~\eref{eqn:spectral_distance} (see \fref{fig:sn_vs_count}). Consequently, while per-sample comparisons could be useful in some cases, we use the more computationally efficient spectral distance metric, \eref{eqn:spectral_distance}, in our experiments. For further details and theoretical motivation, see \sref{apx:ttt_details} and \sref{apx:theory_rr}.

Our experiments show that this procedure virtually never deteriorates performance, as one can always revert to the same radius cutoff used during training (see \sref{sec:experiments}). Additionally, we emphasize that we update the envelope function with the new radius to ensure smoothness of the potential energy surface. We verify that this does maintain energy conservation in \tref{tab:nve_rr}. This refinement method addresses the source of connectivity distribution shifts and serves as an efficient and effective strategy for handling new connectivities.   

\begin{figure*}[t]
     \centering
     \begin{subfigure}[b]{0.6\linewidth}
         \centering
         \includegraphics[width=\textwidth]{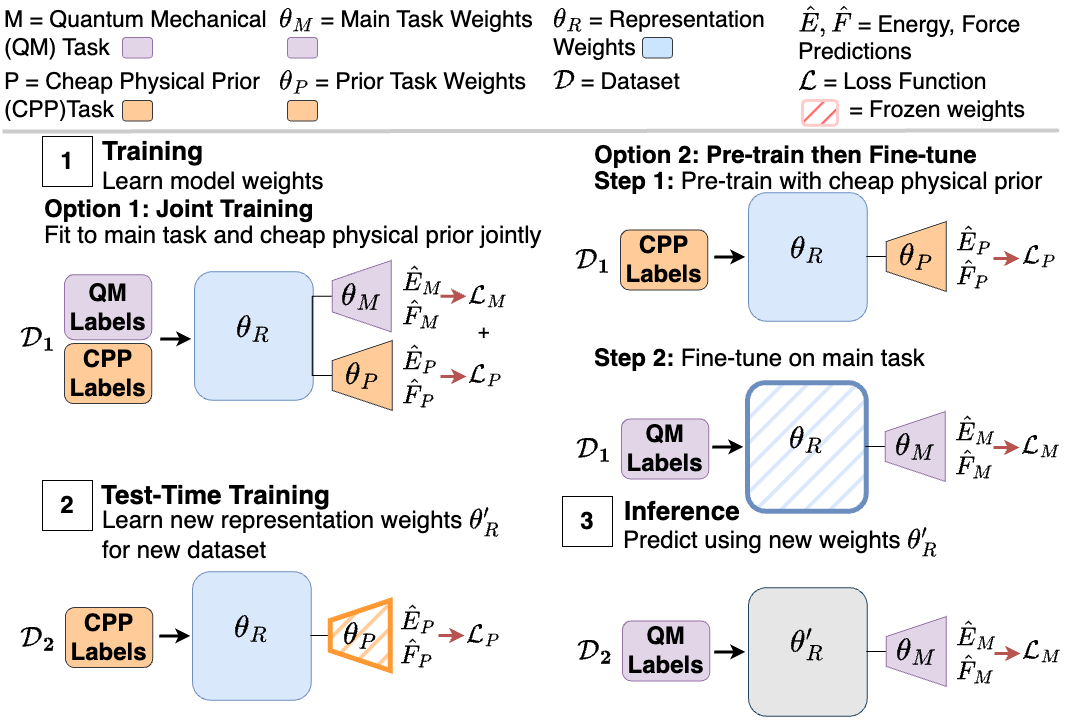}
         \caption{Test-Time Training (TTT)}
         \label{fig:ttt_compressed}
     \end{subfigure}
    \hfill
     \begin{subfigure}[b]{0.3\linewidth}
         \centering
         \includegraphics[height=200pt]{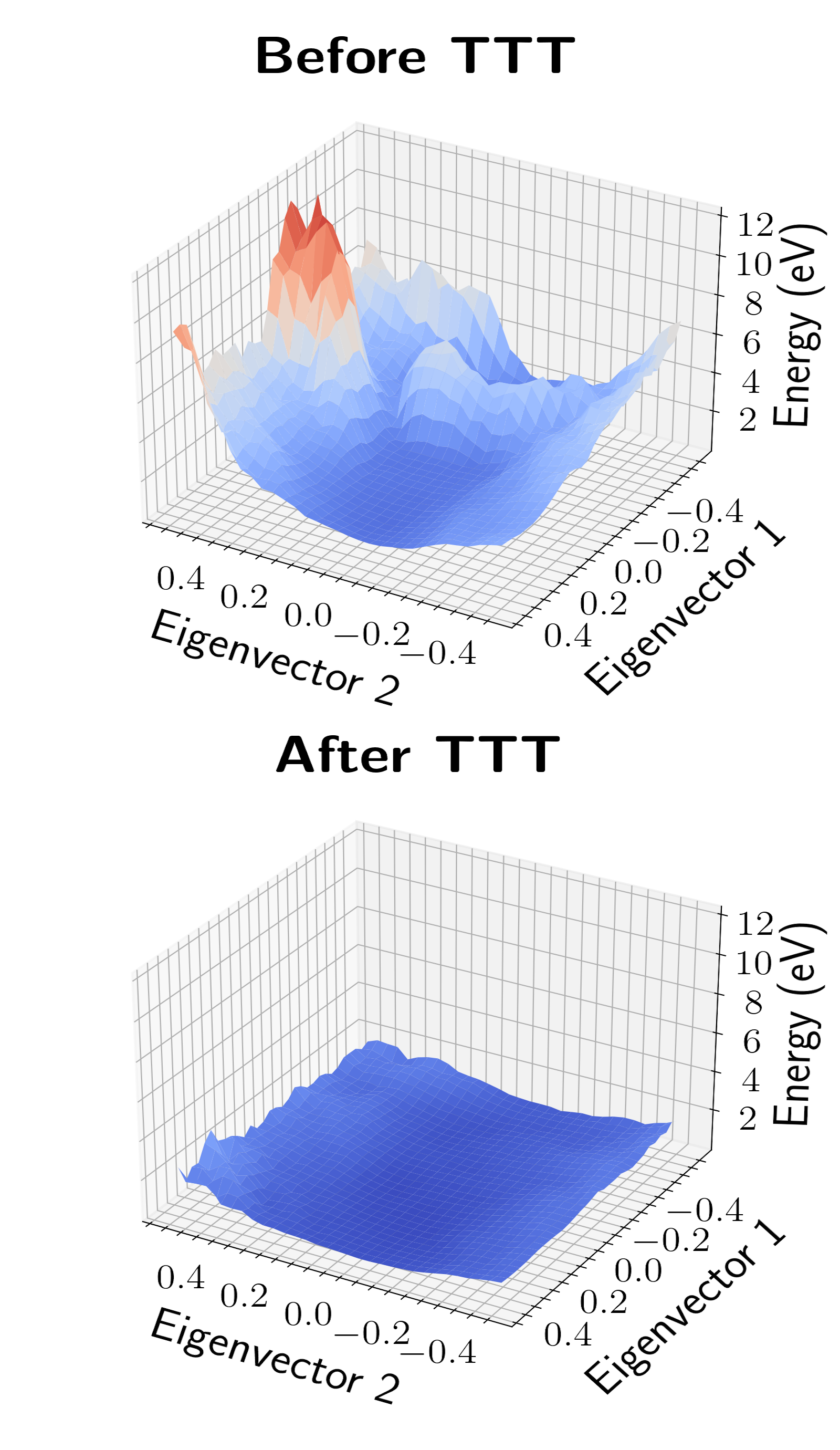}
         \caption{Predicted Potential Energy Surface}
         \label{fig:sali_pes}
    \end{subfigure}
    \caption{\textbf{Test-Time Training Mitigates Distribution Shifts and Smooths Predicted Potential Energy Surfaces.} We hypothesize that due to overfitting, the predicted potential energy surfaces are jagged for out-of-distribution systems. Our proposed test-time training method (TTT, a) regularizes MLFFs by incorporating inductive biases into the model using a cheap prior. Test-time training first learns useful representations from the prior using either joint-training or a pre-train, freeze, and fine-tune approach. TTT then updates the representations at test-time using the prior to improve performance on out-of-distribution samples. We plot the predicted potential energy surface from a GemNet-dT model along the 2 principal components of the Hessian for salicylic acid, a molecule not seen during training, before and after test-time training (b). TTT effectively smooths the potential energy landscape and improves errors.}
    \label{fig:ttt_pes}
    \vspace{-8pt}
\end{figure*}

\vspace{-10pt}
\subsection{Test-Time Training using Cheap Priors}
\label{sec: ttt}
\vspace{-4pt}

We further hypothesize that the current supervised training procedure for MLFFs can lead to overfitting, leading to poor representations for out-of-distribution systems and jagged potential energy landscape predictions (see \fref{fig:sali_pes} for an example on salicylic acid). To address this, we propose introducing inductive biases through improved training and inference strategies to smooth the predicted energy surfaces. The smoother energy landscape from the improved training indicates that the model may have learned more robust representations, mitigating force norm, atomic feature, and connectivity distribution shifts.

We represent these inductive biases as cheap priors, such as classical force fields or simple ML models. These priors can evaluate thousands of structures per second using only a CPU, making them computationally efficient for test-time use. First, we describe our pre-training procedure, which ensures the MLFF learns useful representations from the cheap prior. By leveraging these representations, we can smooth the predicted energy landscape and mitigate distribution shifts by taking gradient steps with our test-time training (TTT) procedure.

\vspace{-8pt}
\paragraph{Pre-Training with Cheap Physical Priors.}

We propose a training strategy that first pre-trains on energy and force targets from a cheap prior and then fine-tunes the model on the ground truth quantum mechanical labels. Our loss function for one structure is defined as:
\begin{equation}
    \label{eqn:loss}
    \mathcal{L}(\mathbf{F}^M, E^M, \mathbf{F^P}, E^P) = \mathcal{L}_M + \mathcal{L}_P = \sum_{l \in \{M, P\}} \left( \lambda_{E^{l}} ||E^l - \hat{E}^l||_2^2 + \lambda_{F^l} \sum_{i=1}^n ||\mathbf{F_i^l} - \mathbf{\hat{F}_i^l}||_2^2 \right),
\end{equation}
where $\hat{E}, \hat{\mathbf{F}}$ are the predicted energy and forces, and $M$ and $P$ denote the main and prior task, respectively. During pre-training, gradient steps are initially only taken on the prior objective, corresponding to $\mathcal{L}_P$. For fine-tuning, the representation parameters, $\theta_R$, learnt from the prior are kept frozen, and the main task parameters, $\theta_M$, are updated by training only on the main task loss, $\mathcal{L}_M$. Pre-training and fine-tuning can also be merged and the model can be \textit{jointly trained} on both the cheap prior targets and the expensive DFT targets (see \fref{fig:ttt_compressed}). This corresponds to training on $\mathcal{L}_P + \mathcal{L}_M$. Freezing or joint-training both force the main task head to rely on features learnt from the prior. This approach acts as a form of regularization, resulting in more robust representations. It enables the prior to be used to improve the features extracted from an out-of-distribution sample at test time, improving main task performance. For more details on the necessity of proper pre-training for test-time training, see \sref{apx:ttt_details}.

\vspace{-6pt}
\paragraph{TTT Implementation Details.} For clarity, let us separate our full model into its three components: $g_{\theta_R}$ (the representation model), $h_{\theta_M}$ (the main task head), and $h_{\theta_P}$ (the prior task head). The representation parameters, $\theta_R$, are learned by minimizing $\mathcal{L}$ during joint training (see \eref{eqn:loss}), or by minimizing $\mathcal{L}_P$ during pre-training and then freezing them during the fine-tuning phase. Test-time training involves the following steps:
\begin{enumerate}
\vspace{-6pt}
    \item \textbf{Updating representation parameters.} At test-time, we update $\theta_R$ by minimizing the prior loss, $\mathcal{L}_P$, on samples from the test distribution $\mathcal{D}_{test}$, which are labeled by the cheap prior. This is expressed as:
\begin{equation}
    \label{eqn:ttt}
    \theta_R^{\prime} = \argmin_{\theta_R} \mathbb{E}_{(\mathbf{r}, \mathbf{z}, \mathbf{F^p}, E^p) \sim \mathcal{D}_{test}} [ \mathcal{L}_P(h_{\theta_P} \circ g_{\theta_R}(\mathbf{r}, \mathbf{z}), \mathbf{F^p}, E^p) ].
\end{equation}
During this process, the prior head parameters, $\theta_P$, are kept frozen during test-time updates. This incorporates inductive biases about the out-of-distribution samples into the model, regularizing the energy landscape and helping the model generalize (see \fref{fig:sali_pes} and \fref{fig: more_pes}).  

\item \textbf{Prediction on test set.} Once the representation parameters are updated, we predict the main task labels for the test set using the newly adjusted representation:
\begin{equation}
    \label{eqn:ttt_pred}
    \hat{E}, \hat{\mathbf{F}} = h_{\theta_M} \circ g_{\theta_R^{\prime}}(\mathbf{r}, \mathbf{z}).
\vspace{-5pt}
\end{equation}
\end{enumerate}
We recalculate the parameters $\theta_R^{\prime}$ with \eref{eqn:ttt} when a new out-of-distribution region is encountered (i.e., when testing on a new system). See \fref{fig:ttt_compressed} for an outline of our method. 

We formalize the intuition behind TTT for MLFFs in the following theorem, where we look at TTT with a simple Lennard-Jones prior \citep{lj}:
\begin{theorem}[] \label{theorem:ttt_mt}
If the reference energy calculations asymptotically go to $\infty$ as pairwise distances go to $0$, then there exist test-time training inputs such that a gradient step on the prior loss, with the Lennard-Jones potential, reduces the main task loss on those inputs. 
\end{theorem}

We prove Theorem \ref{theorem:ttt_mt} by showing that there exist points where the errors on the prior and main task are correlated ($\text{sign}(\hat{E}^P - E^P) = \text{sign}(\hat{E}^M - E^M)$), and that the main task head and the prior task head use similar features ($\theta_P^T \theta_M > 0$). Building off of the theoretical result in \citet{sun2020testtime}, this implies that TTT on these points with prior labels improves main task performance. For a detailed proof, see \sref{apx:theory_rr}.

\vspace{-8pt}
\section{Experiments}
\label{sec:experiments}
\vspace{-5pt}

We conduct experiments on chemical datasets to both identify the presence of distribution shifts and evaluate the effectiveness of our test-time refinement strategies to mitigate these shifts. In \sref{exp:spice_maceoff}, we find distribution shifts on the SPICE dataset with the MACE-OFF foundation model \citep{Eastman2023spice, kovács2023maceoff23}. In \sref{exp:sim_md17}, we explore extreme distribution shifts and demonstrate that our test-time refinement strategy enables stable simulations on new molecules, even when trained on a limited dataset of 3 molecules from the MD17 dataset \citep{chmiela_machine_2017}. Finally, in \sref{exp:h2l_md22}, we assess how our test-time refinement strategy can handle high force norms in the MD22 dataset when the model is trained only on low force norms. Although matching in-distribution performance (without access to ground truth labels) remains a challenging open machine learning problem \citep{sun2020testtime, gandelsman2022testtime}, our experiments indicate that test-time refinement strategies are a promising initial step for addressing distribution shifts with MLFFs. The improvements from these test-time refinement strategies also suggest that MLFFs can be trained to learn more general representations that are resilient to distribution shifts. Additional experiments with more models, datasets, and priors are provided in \sref{apx:more_ttt}.      

\vspace{-8pt}
\subsection{Distribution Shifts: Training on SPICE and testing on SPICEv2}
\vspace{-5pt}

\begin{figure}
    \centering
    \includegraphics[width=0.9\linewidth]{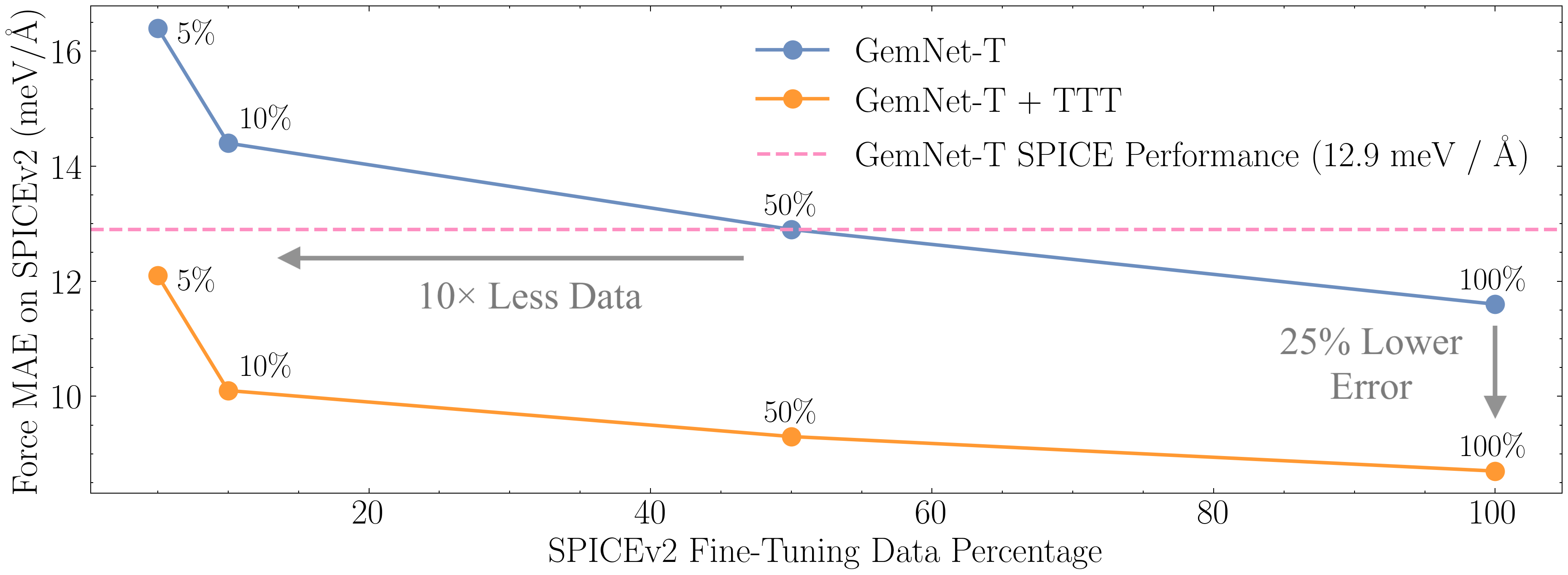}
    \caption{\textbf{Test-Time Training Decreases the Amount of Fine-Tuning Data Needed to Match In-Distribution Performance.} We fine-tune GemNet-T models, trained on SPICE, on new molecules from the SPICEv2 dataset. Applying TTT on the new data before fine-tuning decreases the amount of training data needed to match the in-distribution performance by $10\times$. Applying TTT before fine-tuning also decreases the final error by 25\% when training on all the data.}
    \label{fig:ft_data_spice}
\end{figure}

\begin{figure}
    \centering
    \includegraphics[width=0.8\linewidth]{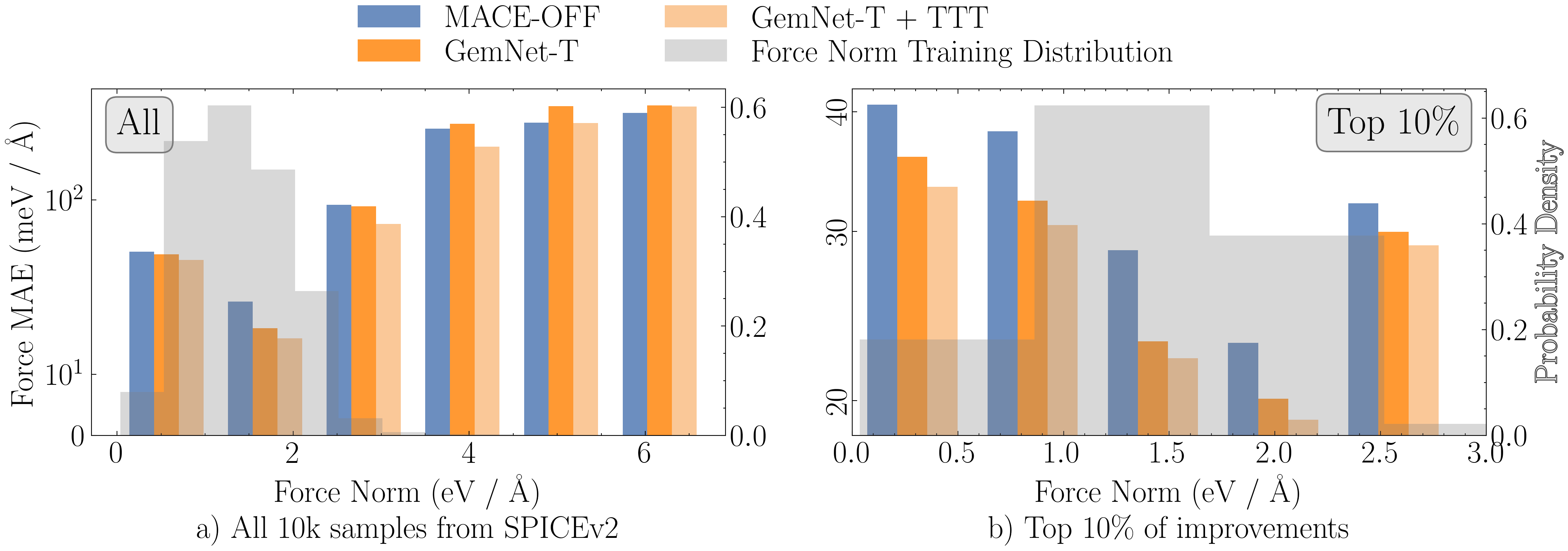}
    \caption{\textbf{Evaluating Distribution Shifts for Force Norms on SPICEv2.} We evaluate MACE-OFF on new molecules from the SPICEv2 dataset with varying force norms. (a) Test structures with different force norms relative to the training distribution (shown in gray) incur larger force errors for MACE-OFF. We also train a GemNet-T model, and then apply test-time training (TTT), mitigating this shift. (b) We highlight the top 10\% of molecules with the greatest improvement to demonstrate that TTT is effective even for structures that are near the training distribution.} 
    \label{fig:spice_force_norms}
\end{figure}

\begin{figure}
    \centering
    \includegraphics[width=0.8\linewidth]{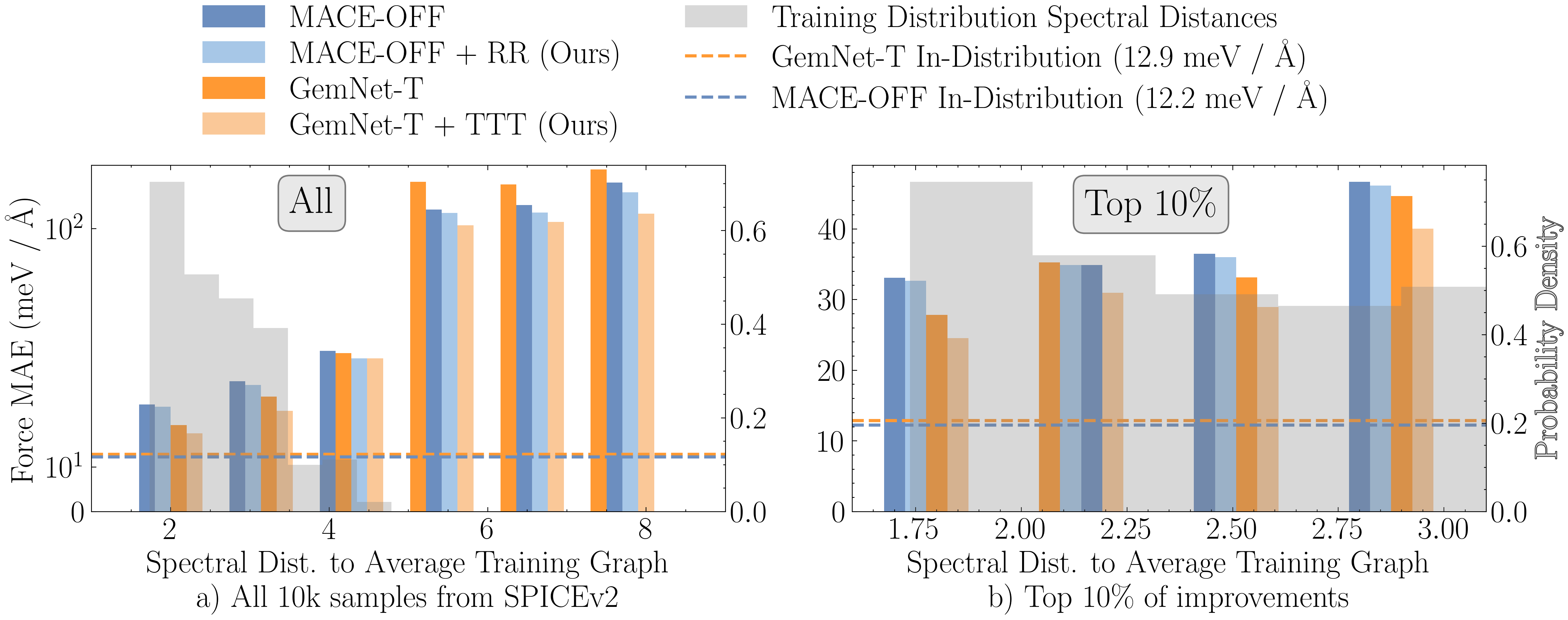}
    \caption{\textbf{Evaluating Connectivity Distribution Shifts on SPICEv2.} We evaluate MACE-OFF on new molecules from the SPICEv2 dataset with varying connectivity, defined by the spectral distance to the average training graph (see \sref{sec: ttrr} for details). (a) Test structures with different connectivity relative to the training distribution (shown in gray) incur larger force errors for MACE-OFF. Test-time training (TTT) applied to a GemNet-T model and test-time radius refinement (RR) applied to MACE-OFF are both able to mitigate this performance drop at minimal computational cost. (b) We highlight the top 10\% of molecules with the greatest improvement to demonstrate that TTT is effective even for connectivities close to the training distribution.}
    \label{fig:spice_connectivity}
\vspace{-12pt}
\end{figure}

\label{exp:spice_maceoff}
We investigate distribution shifts from the SPICE dataset to the SPICEv2 dataset \citep{Eastman2023spice, eastman2024spice2} by analyzing the MACE-OFF foundation model \citep{kovács2023maceoff23}. As shown in \fref{fig:spice_force_norms}, \fref{fig:spice_connectivity}, and \fref{fig:spice_new_elements}, we observe that despite being trained on 951k data points and scaled to 4.7M parameters, MACE-OFF experiences force norm, connectivity, and atomic feature distribution shifts when evaluated on 10k new molecules from SPICEv2 \citep{eastman2024spice2}. Any deviation from the training distribution, shown in gray, predictably results in an increase in force error. 

We evaluate the effectiveness of our test-time refinement strategies in mitigating these distribution shifts. For the MACE-OFF model, we implement test-time radius refinement (RR) by searching over 10 different radius cutoffs and selecting the one that best matches the training Laplacian eigenvalue distribution (see \sref{sec: ttrr}). We also train a GemNet-T model on \update{the same} training data used by MACE-OFF, using the pre-training, freezing and fine-tuning method described in \sref{sec: ttt}, with the sGDML model as the prior \citep{Chmiela_2019sgdml}. To show that TTT is prior agnostic, we additionally train a model that uses the semi-empirical GFN2-xTB as the prior \citep{semiempirical}.
See \ref{apx:exp_details} for more details. 

\vspace{-10pt}
\paragraph{Force Norm Distribution Shifts.}
Both MACE-OFF and GemNet-T deteriorate in performance when encountering systems with force norms different from those seen during training, as shown in \fref{fig:spice_force_norms}. Interestingly, this performance drop occurs for both higher and \textit{lower} force norms than those in the training set. Test-time training reduces errors for GemNet-T on out-of-distribution force norms, and also helps decrease errors for the new systems that are closer to the training distribution. The results in \fref{fig:spice_force_norms} specifically filter out atomic feature shifts and different connectivities to isolate the effect of force norm distribution shifts.

\vspace{-10pt}
\paragraph{Connectivity Distribution Shifts.}
For both MACE-OFF and GemNet-T, force errors increase when the connectivity of a test graph differs from that of the training graphs, as measured by the spectral distance (see \eref{eqn:spectral_distance}). Our test-time radius refinement (RR) technique (see \sref{sec: ttrr}) applied to MACE-OFF effectively mitigates connectivity errors at minimal computational cost. Test-time training also effectively mitigates connectivity distribution shifts, as shown in (\fref{fig:spice_connectivity} and \tref{tab: spice_rr_indiviual}). Note that \fref{fig:spice_connectivity} isolates connectivity distribution shifts by filtering out-of-distribution force norms and atomic features. \update{See \sref{apx:ttrr_jmp} for RR results with the JMP model on the ANI-1x dataset.}  

\vspace{-10pt}
\paragraph{Atomic Feature Distribution Shifts.} MACE-OFF and GemNet-T both perform poorly when encountering molecules with atomic features that differ from their training distributions. In particular, MACE-OFF and GemNet-T struggle with molecules that are both larger and smaller than those seen in training, and with systems that have a different proportion of carbon atoms than seen in training (see \fref{fig:spice_new_elements}). Test-time training reduces errors across both of these atomic feature distribution shifts for GemNet-T. We filter out out-of-distribution connectivities and force norms to isolate the effect of atomic feature distribution shifts.


\vspace{-12pt}
\paragraph{Aggregated Results and Takeaways.} We present aggregated results on the SPICEv2 distribution shift benchmark, where a model is trained on SPICE and evaluated on 10k new molecules from SPICEv2. The large MACE-OFF foundation model trains on 951k samples but still suffers from distribution shifts on the new structures from SPICEv2. We also see that (1) the RR method mitigates connectivity distribution shifts for MACE-OFF at minimal computational cost (see \tref{tab:agg_spicev2}) and
(2) using TTT with the GemNet-T model performs the best on the new molecules from SPICEv2, highlighting the effectiveness of training strategies for mitigating distribution shifts. Practically, these lower force errors also translate into better MD simulations and improved structure relaxations (see \sref{apx:indvidual_mols_spicev2}). 

\FloatBarrier
\noindent
\begin{minipage}{0.45\textwidth}
  Since the improvements from RR and TTT are right-skewed, meaning many molecules show small improvements while some see large gains, we highlight the 10\% of molecules with the greatest improvement in \fref{fig:spice_force_norms}b, \fref{fig:spice_connectivity}b, and \fref{fig:spice_new_elements}b. We also present results for individual molecules in \tref{tab: spice_ttt_indiviual} and \tref{tab: spice_rr_indiviual} to show that TTT and RR can help across a range of errors. Both TTT and RR improve results on molecules that already have low errors, and bring many molecules with high errors close to the in-distribution performance (\update{see \fref{fig: ttt_rr_chem_acc} which shows that more than $8,000/10,000$ molecules have errors below 25 meV / Å}). 

\end{minipage}
\hspace{0.01\textwidth}
\begin{minipage}{0.5\textwidth}
  \setlength{\tabcolsep}{3pt}
\centering
\begin{tabular}{cc}
\toprule
    \textbf{Model}                 & \makecell{\textbf{SPICEv2 Test Set} \\ \textbf{Force MAE (meV/Å)}} \\ \hline
MACE-OFF             & $26.75 \pm 0.65$           \\
 +RR (ours)&$26.0 \pm 0.64$            \\ 
GemNet-T             & $22.9 \pm 1.4$             \\ 
+TTT (ours)& $\mathbf{19.9 \pm 1.0}$    \\ \bottomrule
\end{tabular}
\captionof{table}{\textbf{Aggregated Results on SPICEv2 Distribution Shift Benchmark.} 
We provide aggregated results on the SPICEv2 distribution shift benchmark \update{with 95\% confidence intervals}. TTT and RR are both able to effectively mitigate errors across the $10$k unseen molecules from SPICEv2. The relative improvements observed are in line with previous test-time training work \citep{gandelsman2022testtime, sun2020testtime}.}
\label{tab:agg_spicev2}


\end{minipage}

The ability of TTT and RR to mitigate distribution shifts supports the hypothesis that MLFFs easily overfit to training distributions, even with large datasets. By improving the connectivity and learning more general representations of test molecules, RR and TTT diagnose the specific ways in which MLFFs overfit. These experiments suggest that improved training strategies could help learn more general models.

\paragraph{Test-Time Training and Fine-Tuning.} While TTT can enable accurate MD simulations for new systems without access to any reference labels (see \sref{exp:sim_md17} and \sref{apx:indvidual_mols_spicev2}), a practitioner may prefer to fine-tune a model on more out-of-distribution data to match the in-distribution performance. We examine how TTT can provide a better starting point for fine-tuning by learning more robust representations for new systems. 

We take the GemNet-T models from the previous section and fine-tune them with varying amounts of structures from the SPICEv2 dataset \citep{eastman2024spice2}. We evaluate how much data is required to match the in-distribution force error on SPICE ($12.9$ meV/Å) when tested on the 10k new molecules from SPICEv2. The vanilla GemNet-T model matches the in-distribution performance when trained on half of the SPICEv2 data. In contrast, using our TTT procedure before fine-tuning allows the model to reach the same performance with only $5\%$ of the data — a $10\times$ reduction. Additionally, TTT reduces the final error by $25\%$ even when fine-tuning on the entire SPICEv2 dataset. See \fref{fig:ft_data_spice} for details.

\vspace{-6pt}
\subsection{Evaluating Generalization with Extreme Distribution Shifts: Simulating Unseen Molecules}
\label{exp:sim_md17}

\begin{figure}
\centering
\begin{subfigure}{0.49\linewidth}
  \centering
  \includegraphics[width=0.95\linewidth]{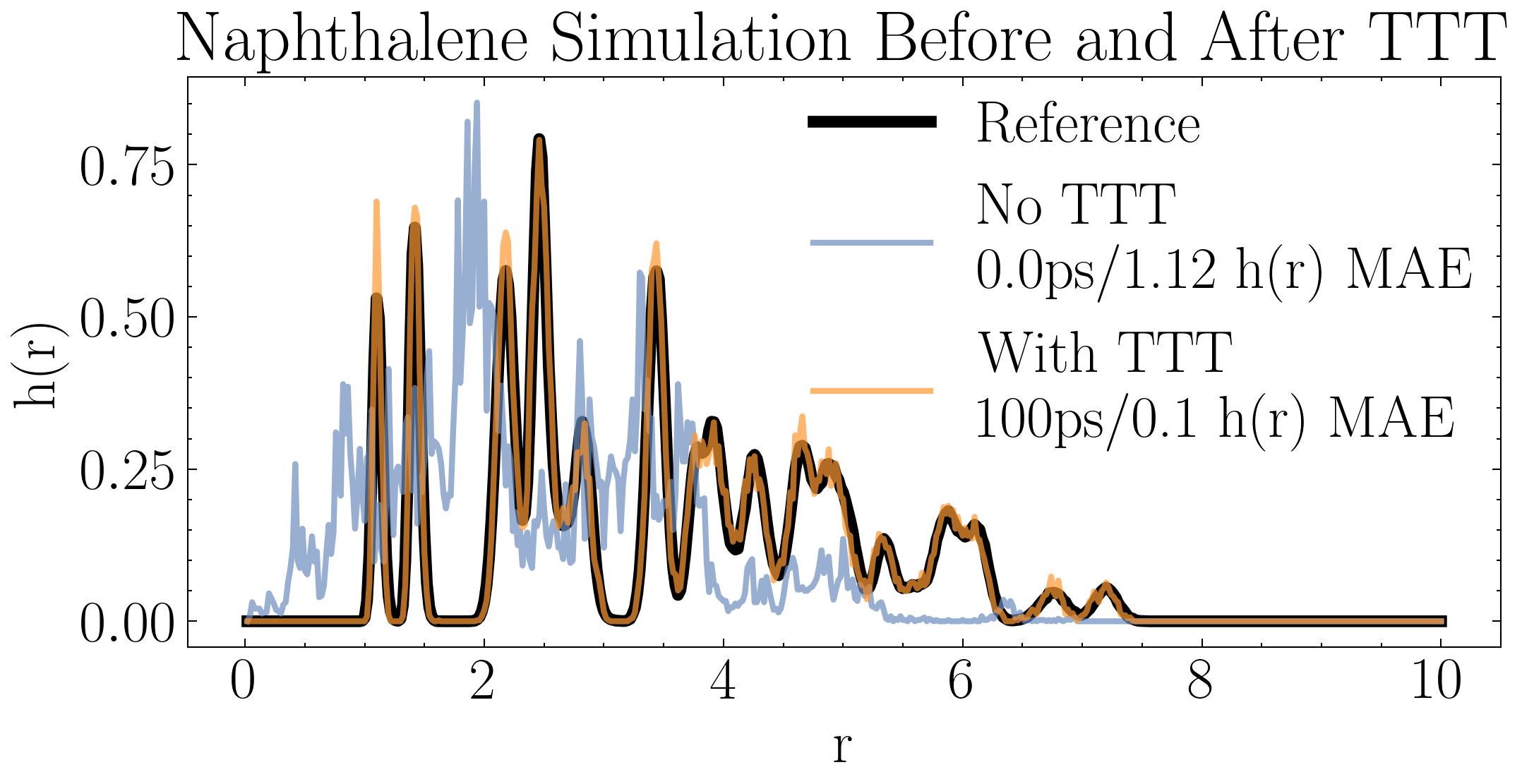}
\end{subfigure}
\begin{subfigure}{0.49\linewidth}
  \centering
  \includegraphics[width=0.95\linewidth]{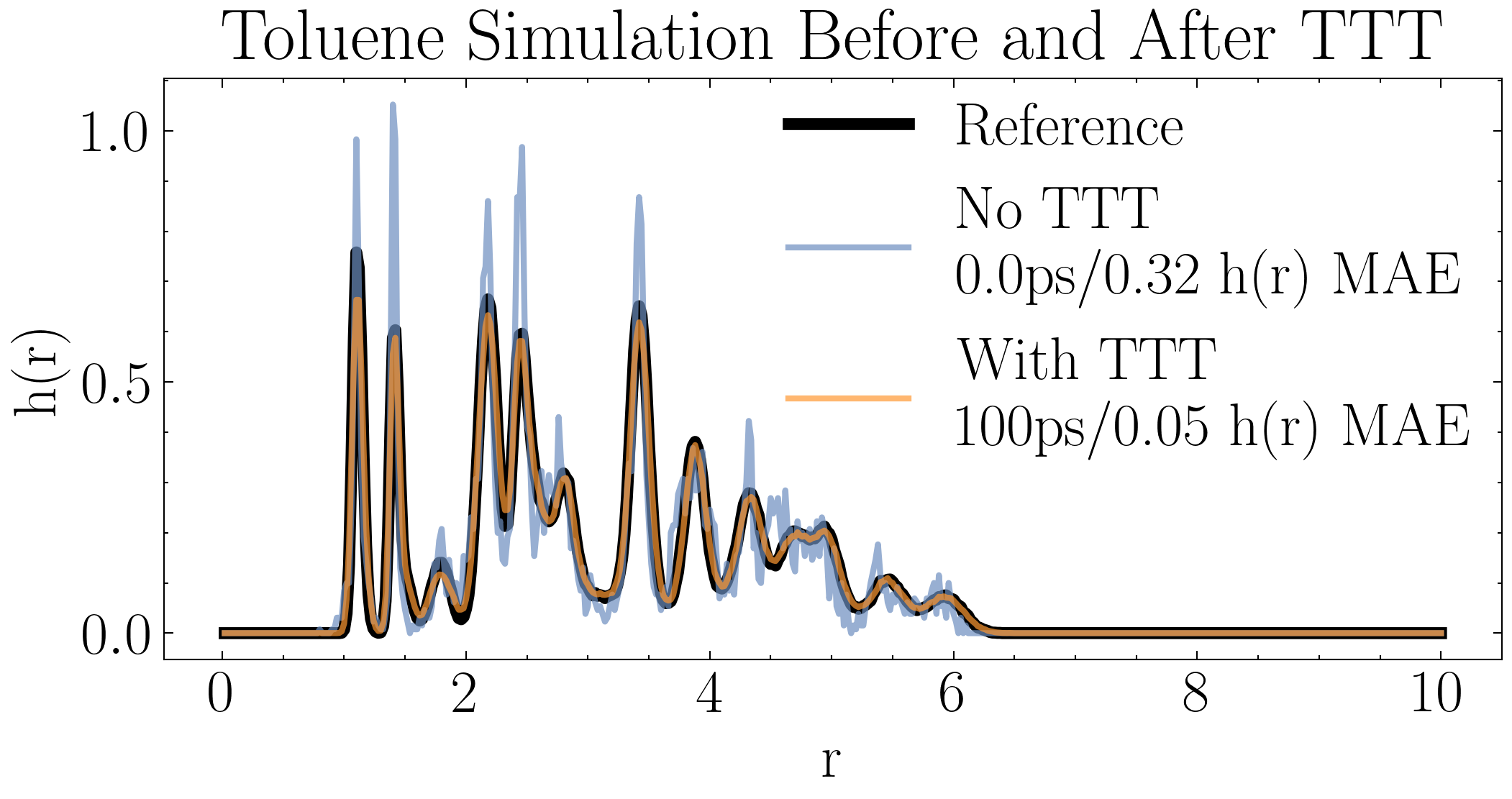}
\end{subfigure}

\caption{\textbf{Testing Molecular Dynamics Simulations.} TTT enables stable simulations that accurately reconstruct observables, such as the distribution of interatomic distances, for molecules not seen during training (orange). In contrast, predictions without TTT for these unseen molecules result in unstable simulations and inaccurate $h(r)$ curves (blue). Simulations without TTT remained unstable even with a timestep reduced by $5,000\times$.} 
\label{fig:hr}
\end{figure}

We establish an extreme distribution shift benchmark to evaluate the generalization ability of MLFFs on the MD17 dataset \citep{chmiela_machine_2017}. This benchmark is specifically designed to highlight how MLFF training strategies tend to overfit to narrow problem settings, and to evaluate how new training strategies can improve robustness. We train a single GemNet-dT model \citep{gasteiger_gemnet_2021} on $10$k samples each of aspirin, benzene, and uracil. We then evaluate whether this model can simulate two new molecules, naphthalane and toluene, which were unseen during training. Next, we evaluate whether TTT can address the distribution shifts to the new molecules. Using the same procedure outlined in \sref{sec: ttt}, we pre-train on the 3 molecules in the training set with the sGDML prior, then freeze the representation model and fine-tune on the quantum mechanical labels. We then perform TTT before simulating the new molecules (see \sref{sec: ttt}). This is an extremely challenging generalization task for MLFFs due to the limited variety of training molecules. Nevertheless, we believe that a model capable of accurately capturing the underlying quantum mechanical laws should be able to generalize to new molecules.   

We evaluate the stability of simulations over time by measuring deviations in bond length, following \citet{fu2023forces}. We additionally calculate the distribution of interatomic distances $h(r)$, a low dimensional descriptor of 3D molecular structures, to measure the quality of the simulations \citep{Zhang_2018, fu2023forces, raja2025stabilityaware}. See \sref{apx:exp_details} for more details.    

\vspace{-10pt}
\paragraph{Simulation Results.} As shown in \fref{fig:hr}, TTT enables stable simulations of unseen molecules that accurately reproduce the distribution of interatomic distances $h(r)$. Without TTT, the GemNet-dT model trained only on aspirin, benzene, and uracil is unable to stably simulate the new molecules and produces poor $h(r)$ curves. Even when we reduce the timestep by a factor of 5,000, the simulations without TTT remains unstable. \update{We also find that TTT enables stable NVE simulations (see \sref{apx:nve_md17}).} Furthermore, TTT provides a better starting point for fine-tuning, decreasing the amount of data needed to reach the in-distribution performance by more than $20\times$ (see \sref{apx:nve_md17}). Given that GemNet-dT + TTT can produce reasonable simulations without access to quantum mechanical labels of the new molecules, test-time refinement methods could be a promising direction for addressing distribution shifts. 

\vspace{-8pt}
\section{Conclusion}

We have demonstrated that state-of-the-art MLFFs, even when trained on large datasets, suffer from predictable performance degradation due to distribution shifts. By identifying shifts in atomic features, force norms, and connectivity, we have developed methods to diagnose the failure modes of MLFFs. Our test-time refinement methods represent initial steps in mitigating these distribution shifts, showing promising results in modeling and simulating systems outside of the training distribution. These results provide insights into how MLFFs overfit, suggesting that while MLFFs are becoming expressive enough to model diverse chemical spaces, they are not being effectively trained to do so. This may indicate that training strategies, alongside data and architecture innovations, will be important in improving MLFFs. Finally, our experiments serve as benchmarks for evaluating the generalization ability of the next generation of MLFFs.

\section{Acknowledgements}

We thank Sanjeev Raja, Rasmus Lindrup, Yossi Gandelsman, Aayush Singh, Alyosha Efros, Eric Qu, and Yuan Chiang for the thoughtful discussions and feedback on this manuscript. This work was supported by the Toyota Research Institute as part of the Synthesis Advanced Research Challenge. This research used resources of the National Energy Research Scientific Computing Center (NERSC), a U.S. Department of Energy Office of Science User Facility located at Lawrence Berkeley National Laboratory, operated under Contract No. DE-AC02-05CH11231. 

\bibliography{example_paper}
\bibliographystyle{icml2025}

\newpage
\appendix
\onecolumn
\section{Details on Test-Time Refinement Training Strategies}
\label{apx:ttt_details}

\subsection{Test-Time Training (TTT)}
We elaborate on the details of our proposed test-time training (TTT) approach.

\paragraph{Model setup.}
Our model consists of the representation model, the main task head, and the prior task head, with parameters $\theta_R$, $\theta_M$, and $\theta_{P}$ respectively:
\begin{enumerate}
    \item The representation model, $\theta_R$, is designed to extract features useful for both the main and prior task heads. These parameters can be trained on both the cheap data from the physical prior and the expensive reference calculations. After pre-training, the representation parameters can be further refined through fine-tuning and test-time training.  
    \item The main task head, $\theta_M$, predicts the energies and forces generated by DFT calculations. This head specifically uses the high-accuracy, expensive quantum mechanical labels produced by DFT for training. 
    \item The prior head, $\theta_P$, predicts the energies and forces from the cheap physical prior, such as classical force fields. This head is trained with the cheap labels produced by the physical prior.
\end{enumerate}
We emphasize that the pre-training and test-time training procedures described in \sref{sec: ttt} are model architecture agnostic. For details on how we split up existing architectures into the representation model, main task head, and prior head, see \sref{apx:exp_details}.

\begin{figure*}[t!]
    \centering
    \begin{subfigure}[t]{0.6\linewidth}
        \centering
        \includegraphics[width=\linewidth]{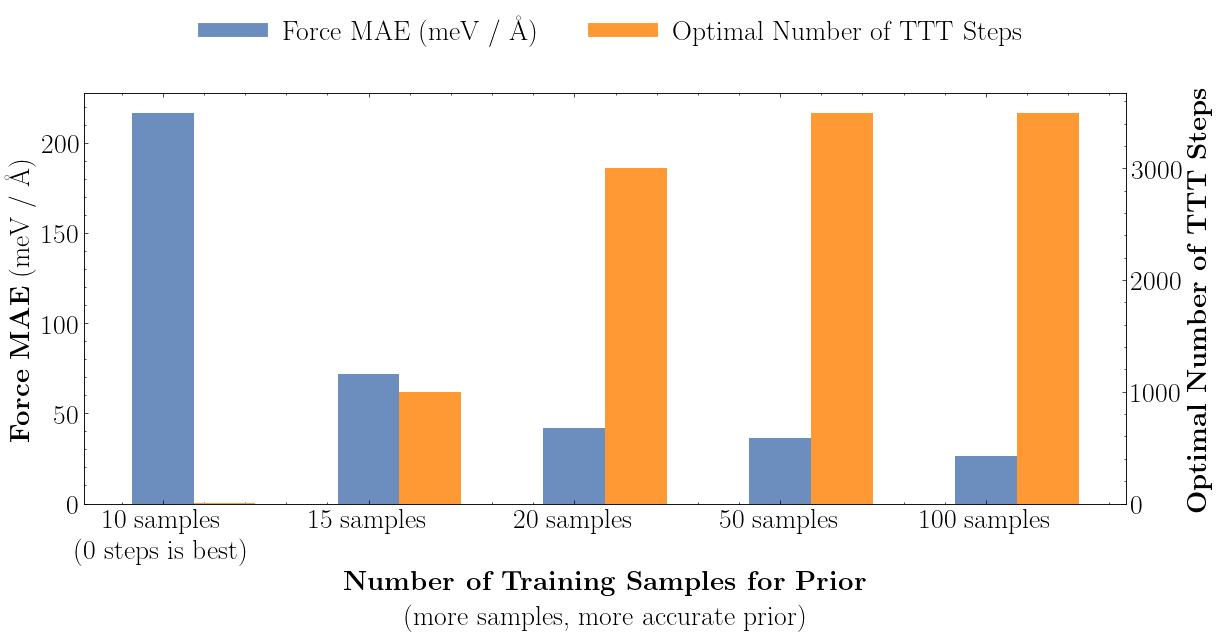}
        \caption{\textbf{Impact of prior accuracy on test-time training (TTT) for naphthalene.} As the prior becomes more accurate by training on more samples, we see larger improvements from TTT (blue bar). This accuracy allows us to take more gradient steps on the prior task (orange bar), without deteriorating performance on the main task. }\label{fig:prior_vs_ttt}
    \end{subfigure}%
    ~ 
    \hfill
    \begin{subfigure}[t]{0.34\linewidth}
        \centering
        \includegraphics[width=\linewidth]{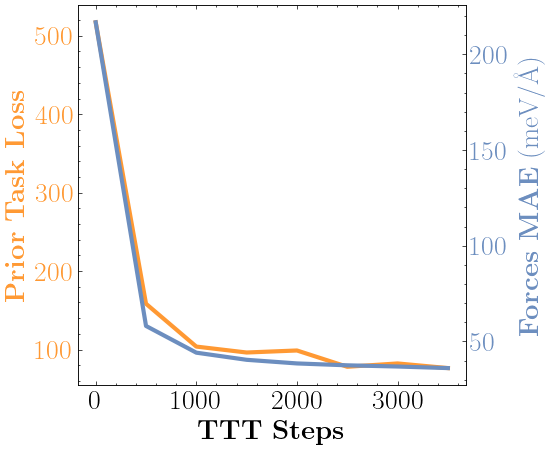}
        \caption{\textbf{Relationship between prior task loss and main task loss.} Fitting to the prior task loss (orange) improves performance on the main task (blue) on naphthalene.}
  \label{fig:ptlmtl}
    \end{subfigure}
    \caption{\textbf{Understanding the Auxiliary Task in TTT.} We train a GemNet-dT model on three molecules from MD17 and perform TTT on naphthalene, a new molecule not seen during training. Our auxiliary objective for TTT is a cheap physical prior. We analyze how the accuracy of the prior affects the performance of TTT (a) and how the prior task loss relates to errors on the main task (b).}
\end{figure*}

\paragraph{Necessity of Proper Pre-training for Test-time Training.} The goal of TTT is to adapt to out-of-distribution test samples using a self-supervised objective at test-time~\citep{sun2020testtime, jang2023testtime, gandelsman2022testtime}. In our case, we use the prior task loss $\mathcal{L}_P$ as the test-time training objective, making the model predict forces and energies labeled by the cheap physical prior. When an out-of-distribution (OOD) sample is encountered at test-time, we can adapt our representation parameters, $\theta_R$, using the prior. This update improves the features extracted from the OOD samples, which in turn smooths the potential energy surface and improves the performance on the main task (see \fref{fig:ptlmtl}). Importantly, naive fine-tuning of the full pre-trained model (both $\theta_R$ and $\theta_M$) hinders the effectiveness of TTT. This is because fine-tuning $\theta_R$ on the main task may cause these parameters to ``forget'' the features learned from the prior during pre-training. If we adjust $\theta_R$ at test-time based solely on the prior targets, this could shift $\theta_R$ away from the representations that $\theta_M$ relies on to make predictions. Thus, for TTT to be successful, it is essential that the main task head depends on the features learned from the prior to make accurate predictions. 

\paragraph{Notes on the Prior.} Although the performance of TTT does improve with a more accurate prior (see \fref{fig:prior_vs_ttt}), we note that even in cases where the prior is poorly correlated with the main task (like with the EMT prior and OC20 in \sref{sec:oc20ttt}), TTT still provides benefits. This is because the prior is only used to learn \textit{representations}, and \textit{not }to directly make predictions on the targets. This means that as long as training on the prior yields good representations, it can be used for TTT.

We also argue that such a prior is in fact widely available. For instance, one could always train an sGDML prior on the existing reference data. Alternatively, one could use a simple potential (like EMT or Lennard-Jones). A different (cheaper) level of quantum mechanical theory can also be used. Alternatively, as with prior TTT work in computer vision, a fully self-supervised objective (like atomic type masking and reconstruction) could also be used. We leave explorations of more priors to future work.

It should be noted that using sGDML as the prior requires a few labeled examples to train the sGDML model for the unseen molecule. We show that as few as 15 labeled examples are sufficient to tune the prior and achieve good TTT results (see \fref{fig:prior_vs_ttt}).  TTT also yields better results than fine-tuning directly on these 15 samples, since the model severely overfits on the small number of samples. We also emphasize that across the board, \textbf{TTT performs better than the prior} (see \tref{tab:acc_prior}).  In addition, the sGDML prior only works on one system, whereas the MLFF can model multiple systems.

\begin{table}
\centering
\begin{tabular}{ll}
\toprule
\textbf{\begin{tabular}[c]{@{}l@{}}Molecule and\\ Number of \\ Training Samples (or source)\end{tabular}} & \textbf{Force MAE} (meV/Å) \\ \toprule
\textbf{Naphthalene}                                                                                   &                            \\
10 samples                                                                                    & 444.03                     \\
15 samples                                                                                    & 123.98                     \\
20 samples                                                                                    & 51.77                      \\
50 samples                                                                                    & 42.28                      \\
100 samples                                                                                   & 20.86                      \\ \hline
 \textbf{Toluene}&\\
 50 samples&44.82\\ \hline
\textbf{Ac-Ala3-NHMe}&                            \\
\citep{chmiela_accurate_2023}                                                                                   & 34.25\\ \hline
 \textbf{Stachyose}&\\
 \citep{chmiela_accurate_2023}&29.05\\ \hline
\textbf{Buckyball Catcher}&                            \\
100 samples                                                               & 99.15                      \\ \hline
\textbf{Average over 10k molecules from SPICEv2}&                            \\
$\sim20$ samples& 62.25 (up to 724.5)\\ \hline
 \textbf{EMT}&\\
 \citep{JACOBSEN_emt} &415\\ \hline
 \textbf{GFN2-xTB on SPICEv2}&\\
 \citep{semiempirical} & 201.6 \\ \bottomrule
\end{tabular}
\caption{\textbf{Accuracy of Prior for TTT.} TTT always outperforms the prior.}
\label{tab:acc_prior}
\end{table}

\paragraph{Limitations.} \update{Test-time training incurs extra computational cost, mainly due to the gradient steps taken at test time. This cost is negligible compared to the overall training time of a model, and negligible compared to the time it takes to run simulation with the model. Additionally, our instantiation of TTT requires access to a prior. However, a suitable prior is almost always available since one can always use a widely applicable analytical or semi-empirical potential.} 

\subsection{Test-Time Radius Refinement (RR)}

In this section we discuss further details about our RR approach (for theoretical justification, see \sref{apx:theory_rr}). Although one potential worry about using RR is that it might introduce potential discontinuities, we emphasize that we update the envelope function to ensure that the predicted potential energy surface remains smooth with the new radius. When running MD simulations, we choose the updated radius at the beginning and keep it fixed over the course of simulation. We verify that this maintains a conservative force field by running NVE simulations (see \tref{tab:nve_rr}). Additionally, one might worry that the introduction of new edges will cause the model to overcount certain interactions. However, since edge features contain distance information, and since the model is trained on structures with varied edge distances, a well-trained model should be able to extract features from different edges. We note again that this is not an issue inherent to RR, since GNN-based MLFFs already deal with atoms entering a neighborhood during the course of simulation. Empirically, our experiments show that RR decreases force errors and improves simulation stability (see \sref{exp:spice_maceoff} and \tref{tab: spice_rr_indiviual}).
  
\begin{table}[]
\centering
\begin{tabular}{ll}
\toprule
Model                        & Energy Deviation (eV) \\ \hline
MACE-OFF                     & 0.0036 ± 0.0004       \\
MACE-OFF + RR                & 0.0049 ± 0.0022       \\
GemNet-dT (non-conservative) & \textgreater{}1.0     \\ \bottomrule
\end{tabular}
\caption{\textbf{RR Maintains Energy Conservation.} When using RR, we select the updated radius for the new molecule at the start of simulation and then keep it fixed. We update the envelope function to ensure smoothness of the predicted potential energy surface with the new radius. We run 10 ps NVE simulations to verify that RR does maintain a conservative force field.}
\label{tab:nve_rr}
\end{table}

\section{Theoretical Motivation for Test-Time Refinement}
\label{apx:theory_rr}

\paragraph{Test-Time Training.} We provide theoretical justification for the intuition behind test-time training for machine learning force fields: if we have access to a cheap prior that approximates the reference labels, then taking gradient steps on the prior task will improve performance on the main task. Although making rigorous theoretical statements about deep neural networks in general is challenging, following previous test-time training works \citep{sun2020testtime}, we assume a linear model to provide theoretical guarantees. 

\begin{theorem}[TTT with a Lennard-Jones Prior Improves Performance on Quantum Mechanical Predictions]
\label{theorem:ttt}

Consider the linear model with representation parameters $R \in \R^{f \times d}$ , main task head parameters $m \in \R^{d \times 1}$ and prior task head parameters $p \in \R^{d \times 1}$. Main and prior task head predictions on input $x \in  \R^{f \times 1}$ are given by $\hat{E}^P = x^TRp, \; \hat{E}^M = x^TRm$. Let $R^{\prime}_x$ be the updated representation weight matrix after one step of gradient descent on the prior loss with $x$ as input, and learning rate $\eta$, and energy labels given by the Lennard-Jones potential:
$$R^{\prime}_x \leftarrow R - \eta \nabla_R \mathcal{L}_P(x^TRp, E^P) = R - \eta (E^P - x^T R p)(-xp^T).$$
If the reference energy calculations asymptotically go to $\infty$ as pairwise distances go to $0$, and the features are chosen such that the activations ($A = XR$) have column rank $d$, then there exist inputs $x$ such that:
$$\mathcal{L}_M(x^TR^{\prime}_xm, E^{M}) < \mathcal{L}_M(x^TRm, E^{M}).$$
In other words, taking gradient steps on the prior reduces the main task loss.    
\end{theorem}

The proof builds on the main theoretical result presented by \citet{sun2020testtime}:

\begin{proof}
Based on \citet{sun2020testtime}, it suffices to show that there exist inputs $x$ such that:
\begin{equation}
    \label{eqn: cond_sign}
    \text{sign}(E^P - x^T R p) = \text{sign}(E^M - x^T R m),
\end{equation}
and
\begin{equation}
    \label{eqn: cond_inner}
    p^T m > 0.
\end{equation}
In other words, the errors are correlated, and the task heads use similar features. 

To see that there exist test points where the errors are correlated (\eref{eqn: cond_sign}), we use the fact that both the Lennard-Jones prior and the reference energies (by assumption) go asymptotically to $\infty$ as pairwise distances go to $0$. Our linear model, however, can only make predictions within a bounded range over a bounded domain. Therefore, there clearly exists some $x$ with pairwise distances small enough such that
$$x^TAp < E^P \text{ and } x^TAm < E^M,$$ 
implying that 
$$(E^P - x^T A p), (E^M - x^T A m) > 0.$$ 

In other words, we can always find points where our model will underpredict both the prior and the main task energies. 

To see that the task heads use similar features (\eref{eqn: cond_inner}), we consider a set $X \in \R^{n \times f}$ of $n$ training examples. If we freeze the representation parameters as described in \sref{sec: ttt}, then by least squares the learned $p$ and $m$ are:
$$p = (A^TA)^{-1}A^T y^P, \; m = (A^TA)^{-1}A^T y^M$$
where $y^P, y^M$ are the vectors of prior and main task energies, respectively. Then:
\begin{equation}
    \label{eqn: pm_inner}
    p^T m = (y^P)^T A ((A^T A)^{-1})^T (A^TA)^{-1}A^T y^M = (y^P)^T C y^M.
\end{equation}

By the assumptions, we can express $y^P, y^M$ in the orthogonal eigenbasis of C (with eigenvalues and eigenvectors $\lambda_i$, $v_i$):
$$y^P = \sum_j c_j v_j, \; y^M = \sum_k c_k v_k$$
Since we can always choose test-time training inputs where both the prior and the reference energy goes to $\infty$, then there clearly exist points where:
\begin{equation}
    \label{eqn: y_corr}
    (y^P)^T y^M > 0,
\end{equation}
implying that $y^P, y^M$ share a common eigenvector with $c_j c_k > 0$.

Returning to \eref{eqn: pm_inner}:
$$(y^P)^T C y^M = (\sum_j c_j v_j^T) C (\sum_k c_k v_k) =$$
$$(\sum_j c_j v_j^T) (\sum_k \lambda_k c_k v_k) > 0$$where the last inequality holds because of \eref{eqn: y_corr} and the fact that C is positive definite.
 
 To summarize, since the prior approximates the reference energies, we have shown we can find points where the errors are correlated and the model uses the same features. Using the theorem from \citet{sun2020testtime}, this implies that gradient steps on the prior task improve performance ont he main task, concluding the proof. 
\end{proof}

\begin{figure}[H]
    \begin{subfigure}{0.68\textwidth}
    \centering
    \includegraphics[width=\linewidth]{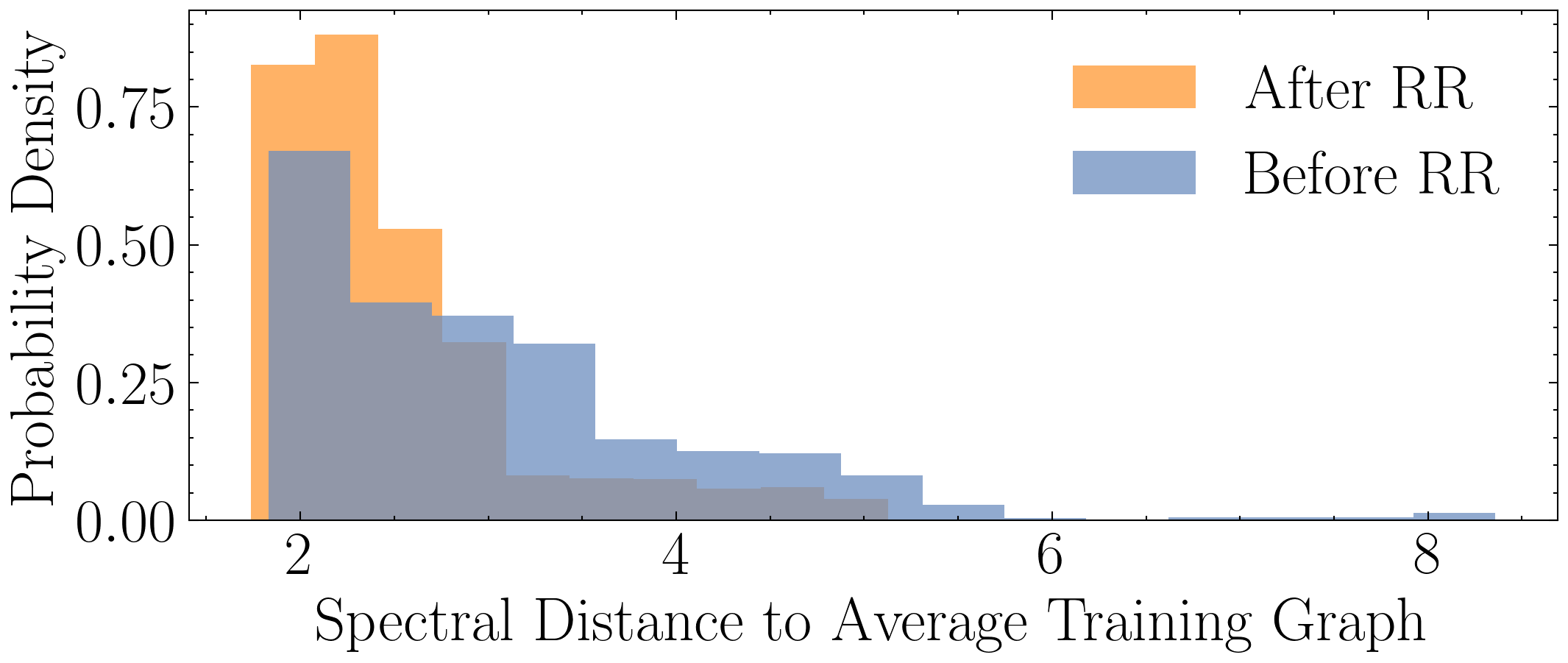}    
    \caption{Distance to Training Distribution}
    \end{subfigure}
    \begin{subfigure}{0.3\textwidth}
    \centering
    \includegraphics[width=0.84\linewidth]{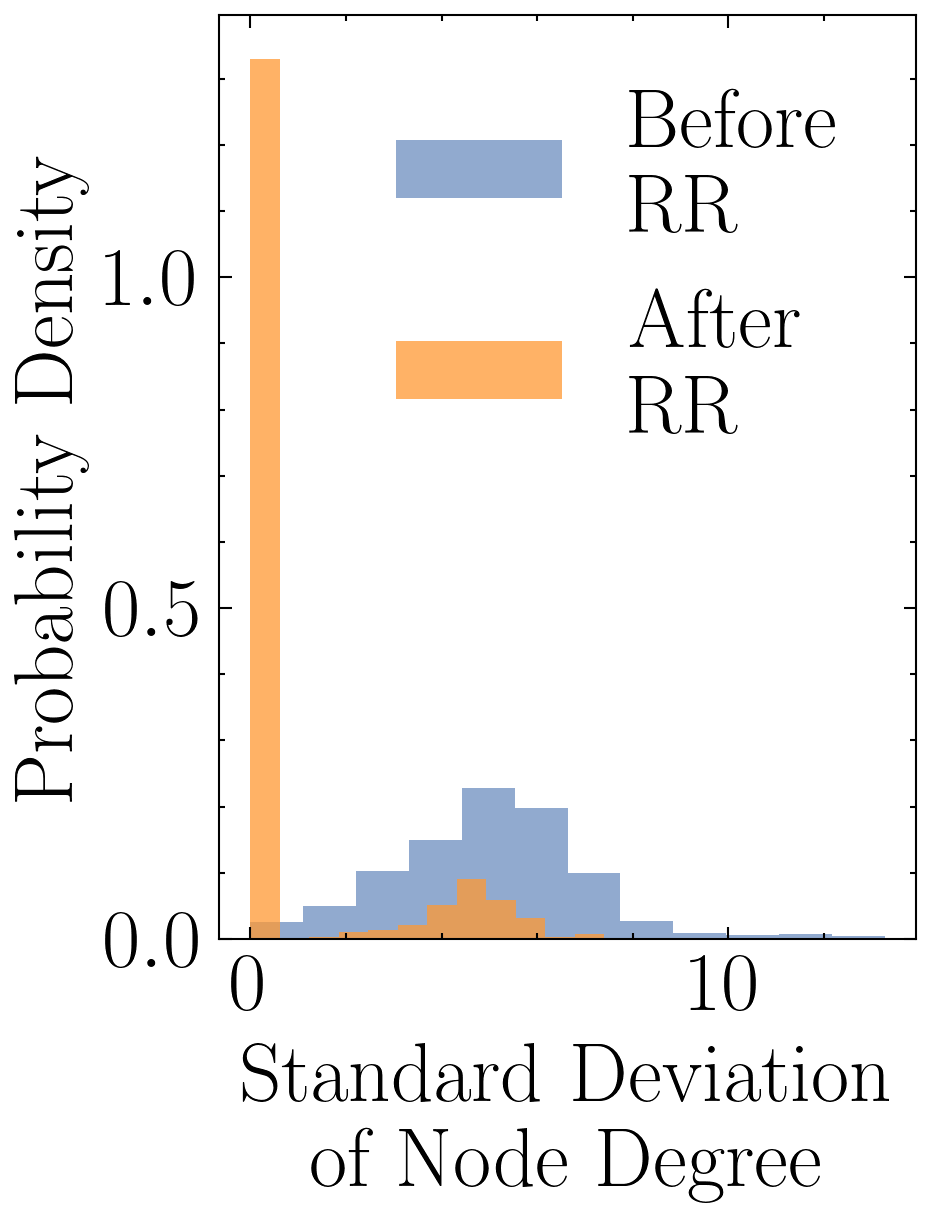}
    \caption{Node Degree}
    \end{subfigure}
    \caption{\update{\textbf{Effect of Radius Refinement (RR) on Molecular Graph Connectivities.} We compare the connectivities of new molecular systems from the SPICEv2 dataset to the training distribution from SPICE, using the MACE-OFF training radius cutoff.  Our results show that RR brings the connectivities of these molecular systems closer to the training distribution, as measured by the spectral distance (a) (note that for some molecular systems, the connectivity doesn't change unless the radius is made very small). Additionally, RR leads to more regular graph structures, with a reduced standard deviation of node degrees (b), indicating that the graphs are more regular.
    }}
    \label{fig:conn_rr}
\end{figure}

\paragraph{Test-Time Radius Refinement.}
\update{Our test-time radius refinement strategy is based on the theoretical finding presented by \citet{bechlerspeicher2024gnnregular}, which states that GNNs tend to overfit to generally regular and well-connected training graphs. Although the theorems are presented for classification problems, they provide intuition and motivation for our RR approach. We restate some of the important theoretical results here (for the proofs and more details see \citet{bechlerspeicher2024gnnregular} and \citet{gunasekar2019implicitbiasgradientdescent}).}

\update{\begin{theorem}
[Extrapolation to new graphs \citep{bechlerspeicher2024gnnregular}]
\label{theorem:ext_new_graphs}
    Let $f^*$ be a graph-less target function (it does not use a graph to calculate its output). In other words, $f^*(X,A) = f^*(X)$, where $X$ are node features and A is the adjacency matrix of a graph. There exist graph distributions $P_1$ and $P_2$, with node features drawn from the same fixed distribution, such that when learning a linear GNN with gradient descent on infinite data drawn from $P_1$ and labeled with $f^*$, the test error on $P_2$ labeled with $f^*$ will be $\geq \frac{1}{4}$. In other words, the model fails to extrapolate to the new graph structures at test time.
\end{theorem}}

\update{Mapping this to MLFFs, theorem \ref{theorem:ext_new_graphs} suggests that a GNN trained on specific types of molecular structures (i.e., acyclic molecules) could fail to generalize to new connectivities at test time (i.e., a benzene ring).}

\update{\begin{theorem}[Extrapolation within regular graph distributions \citep{bechlerspeicher2024gnnregular}]
Let $D_G$ be a distribution over $r$-regular graphs and $D_X$ be a distribution over node features. A model trained on infinite samples from $D_G, D_X$ and labeled by a graph-less target function $f^*$ will have zero test error on samples drawn from $D_X, D_{G^{\prime}}$ (and labeled by $f^*$), where $D_{G^{\prime}}$ is a distribution over $r^{\prime}$-regular graphs.
\end{theorem}}
\update{In other words, generalizing across different types of regular graphs is easier for GNNs. Based on these theorems and our observation that many molecular datasets (MD17, MD22, SPICE) contain generally regular and well-connected graphs, we are motivated to find ways to make testing graphs look more like the training distribution (generally regular and well-connected) to help the models generalize. The observation that graphs for MLFFs are often generated by a radius cutoff led us to develop the RR method presented in \sref{sec: ttrr}. See \fref{fig:conn_rr}, which empirically shows that RR makes graphs more regular and brings them closer to the distribution of training connectivities, aligning with our theoretical intuition. While we think it is an interesting direction for future research to continue exploring the theoretical properties of graph structure distribution shifts.}

\section{Additional Test-Time Refinement Results}
\label{apx:more_ttt}

\begin{figure}
    \centering
    \includegraphics[width=0.8\linewidth]{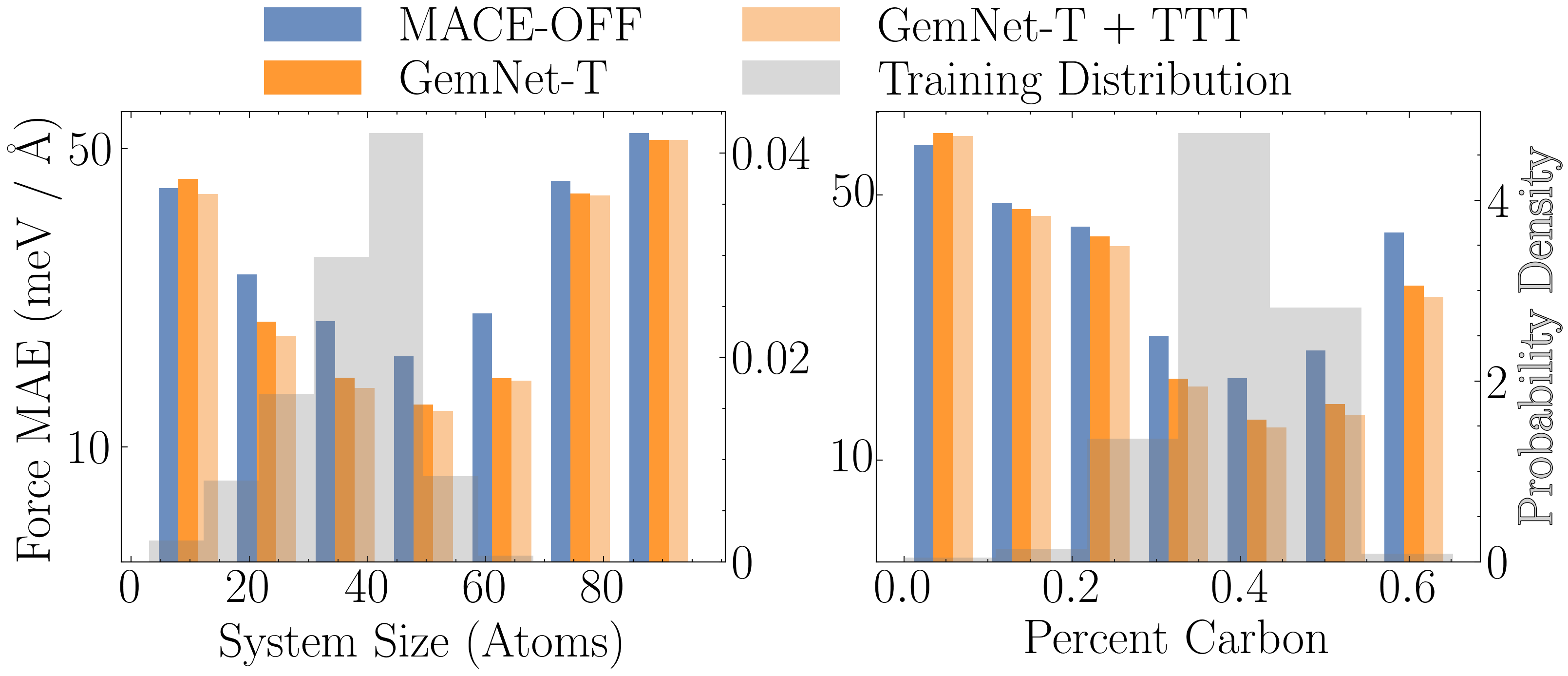}
    \caption{\textbf{Assessing the Impact of Atomic Feature Distribution Shifts on Model Performance on SPICEv2 Benchmark} We evaluate models trained on 951k samples from SPICE on new molecules from SPICEv2. The MACE-OFF model deteriorates in performance when encountering systems of different sizes or molecules with a different proportion of carbons atoms compared to its training set. We train a GemNet-T model on the 951k samples and run TTT---this is able to mitigate the atomic feature distribution shifts.}
    \label{fig:spice_new_elements}
\end{figure}

We provide additional test-time refinement experiments using more models, datasets, and priors. Although these constitute challenging generalization tasks, test-time refinement shows promising first steps at mitigating distribution shifts and generalizing to new types of systems.

\subsection{Further Results on SPICEv2 Distribution Shift Benchmark}
\label{apx:indvidual_mols_spicev2}
Since the TTT and RR results for the SPICEv2 distribution shift benchmark (see \sref{exp:spice_maceoff}) are right skewed, there are many molecules that only improve slightly and a few that improve dramatically. \update{In \tref{tab: spice_ttt_indiviual} and \tref{tab: spice_rr_indiviual}, we highlight results from 6 randomly selected molecules from the top 1,000 most improved with TTT and RR. Specifically, two molecules were randomly chosen from each of the following force error bins: $0–40, 40–100,$ and $>100$ meV / Å.} These results show that TTT and RR help across a range of errors: bringing high errors down to below $40$ meV / Å, and improving results on already low errors.

\begin{table}[h!]
\centering
\setlength{\tabcolsep}{0.2pt} 
\renewcommand{\arraystretch}{1.3}
\begin{tabular}{lcccccc}
\toprule
 & \textbf{$C_4NH_{12}$} & \textbf{$N_3C_5H_3$} & \textbf{$IC_2H$} & \textbf{$ClOC_{14}NH_{15}$} & \textbf{$C_{10}N_2C_3H_{14}$} & \textbf{$O_3P$} \\ \hline
\textbf{GemNet-T} & 28 & 18 & 93 & 55 & 210 & 748 \\ 
Force MAE (meV/Å) / Stability (ps)& $100\pm 0$& $100\pm 0$& $14.7 \pm 1.2$& $100\pm0$& $100\pm0$& $18.5 \pm 0.7$\\ \hline
\textbf{GemNet-T + TTT} & 16 & 13 & 42 & 31 & 70 & 91 \\ 
Force MAE (meV/Å) / Stability (ps)& $100\pm 0$& $100\pm 0$& $38.2\pm 6.0$& $100\pm0$& $100\pm0$& $100\pm 0$\\ \bottomrule
\end{tabular}
\caption{\textbf{Benefit of Test-Time Training (TTT).} We evaluate a GemNet-T model trained on 951k samples from SPICE on 10k new molecules from SPICEv2. We highlight specific examples from SPICEv2 where TTT provides large improvements. TTT can decrease errors by an order of magnitude, and can bring errors close to in-distribution performance. Even when errors are already low, TTT can further reduce errors. \update{TTT also improves NVT simulation stability (mean $\pm$ standard deviation
reported over 3 seeds).}}
\label{tab: spice_ttt_indiviual}
\end{table}

\begin{table}[h!]
\centering
\setlength{\tabcolsep}{2pt} 
\renewcommand{\arraystretch}{1.3}
\begin{tabular}{lcccccc}
\toprule
 & \textbf{$IC_2H$} & \textbf{$O_5N_3C_{16}H_{35}$} & \textbf{$N_4C_7H_{11}$} & \textbf{$O_4C_2PH_6$} & \textbf{$C_6N_2H_{12}$} & \textbf{$SC_6H_4$}\\ \hline
\textbf{MACE-OFF} & 23 /& 12 / & 58 /& 79 /& 875 /& 109 /\\ 
Force MAE (meV/Å) / Stability (ps)& $100 \pm 0$& $38.7 \pm 12.6$& $100 \pm 0$& $100 \pm 0$& $62.8 \pm 26.3$& $100 \pm 0$\\ \hline
\textbf{MACE-OFF + RR} & 16 / & 9 /& 39 / & 49 /& 374 /& 69 / \\ 
Force MAE (meV/Å) / Stability (ps)& $ 100 \pm 0$& $78.9 \pm 16.3$& $100 \pm 0$& $100 \pm 0$& $100 \pm 0$& $100 \pm 0$\\ \bottomrule
\end{tabular}
\caption{\textbf{Benefit of Radius Refinement (RR).} We evaluate MACE-OFF, trained on 951k samples from SPICE, on 10k new molecules from SPICEv2. We highlight specific molecules from SPICEv2 to show that RR improves errors across a range of values. \update{RR also improves NVT simulation stability (mean $\pm$ standard deviation reported over 3 seeds).}}
\label{tab: spice_rr_indiviual}
\end{table}

\begin{figure}
\centering
\begin{subfigure}{0.45\textwidth}
  \centering
\includegraphics[width=\linewidth]{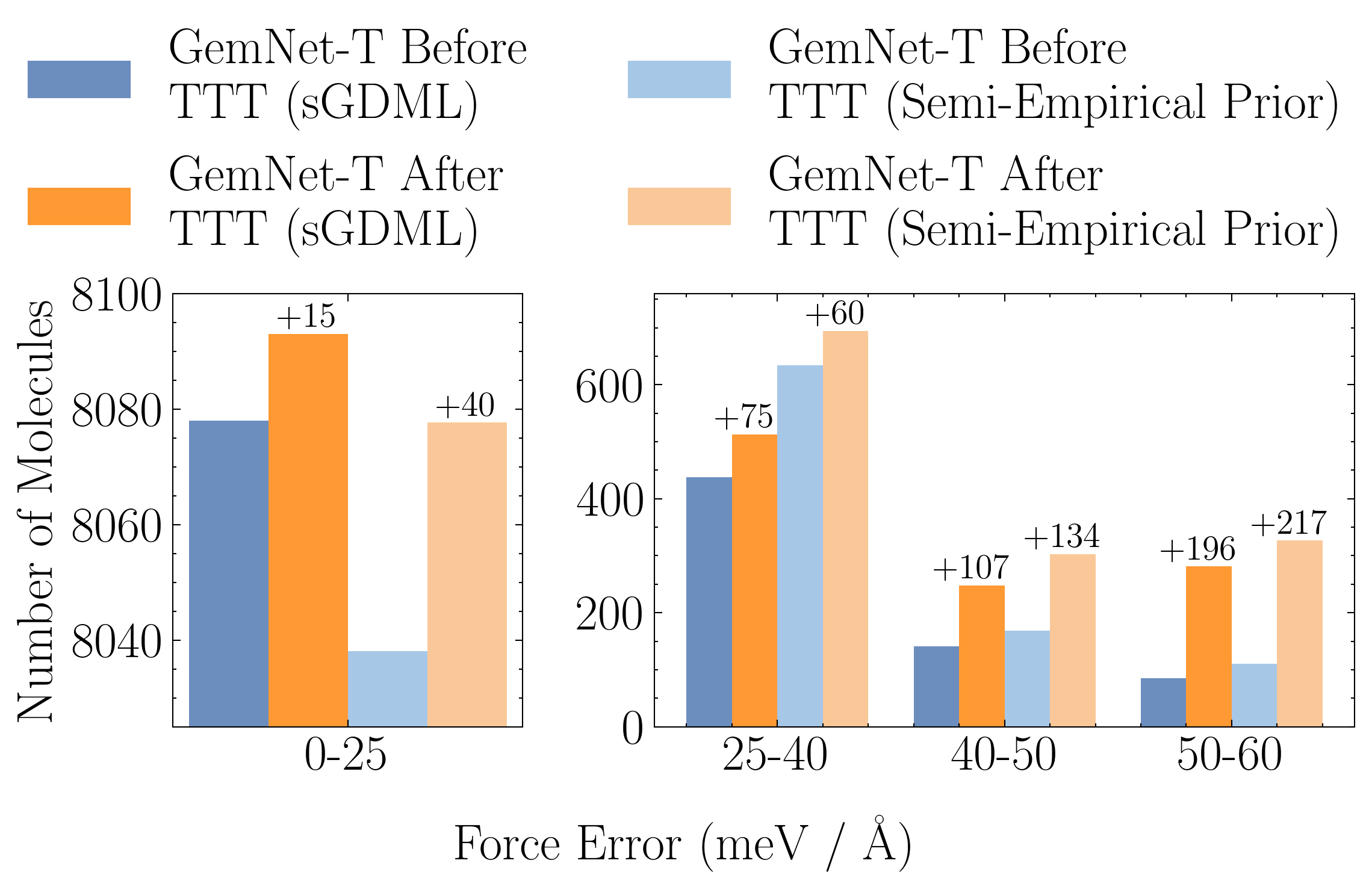}
  
\end{subfigure} 
\begin{subfigure}{0.45\textwidth}
  \centering
\includegraphics[width=\linewidth]{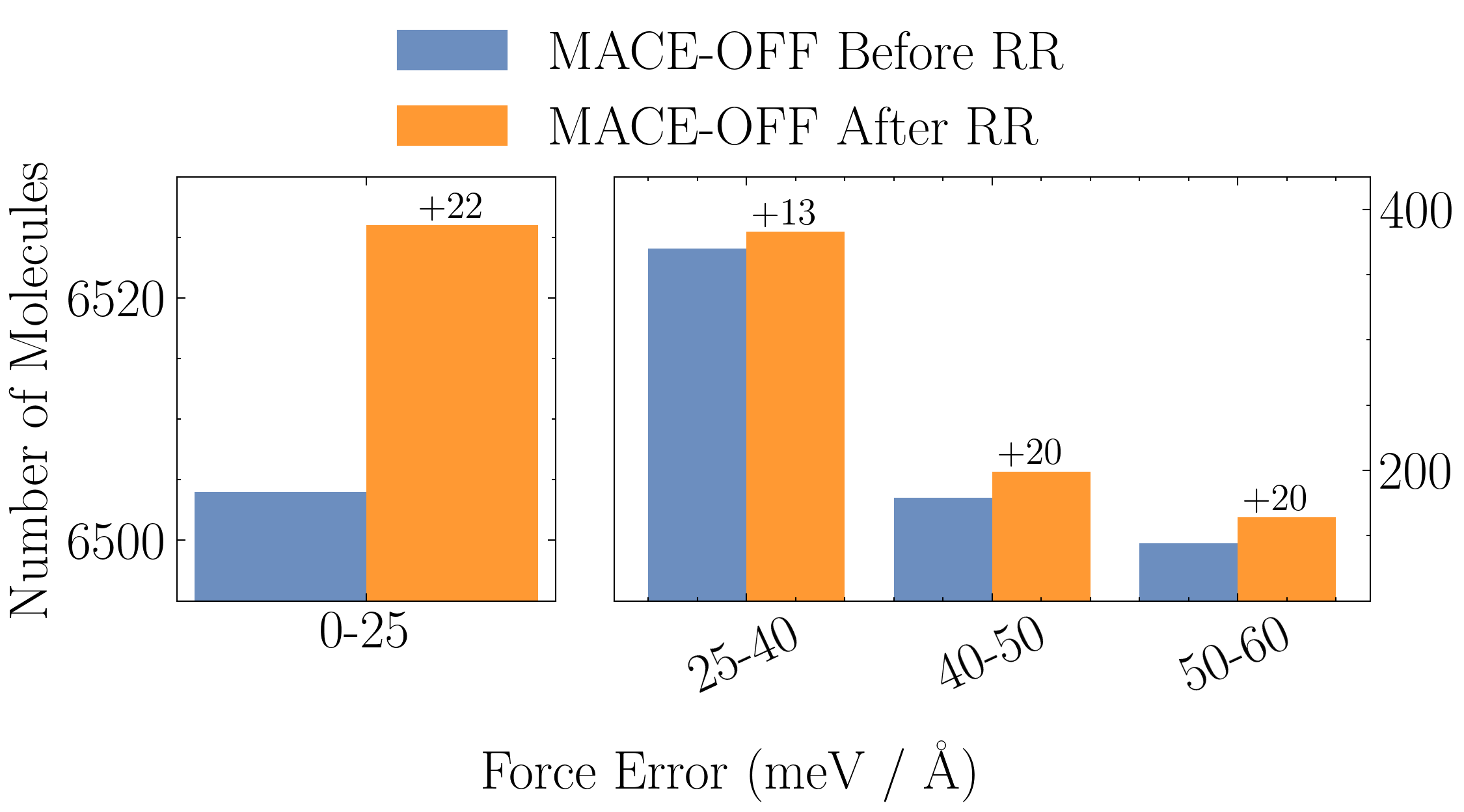}
  
\end{subfigure}
\caption{\update{\textbf{Test-Time Training and Radius Refinement Strategies for Improved Molecular Force Prediction.} 
We train a GemNet-T model (left) on 951k samples from the SPICE dataset and evaluate it on new molecules from the SPICEv2 dataset. We also evaluate the MACE-OFF model (right), which was also trained on the same 951k samples from SPICE. We plot the number of molecules that fall into specific force error bins to show that TTT (\textbf{left}) and RR (\textbf{right}) help improve errors for hundreds of molecular systems. As with previous test-time training works, improvements are more challenging to achieve for systems with lower initial errors (i.e., those closer to in-distribution performance), but TTT and RR still help bridge the gap to in-distribution performance.}}
\label{fig: ttt_rr_chem_acc}
\end{figure}

\update{We also explicitly quantify in \fref{fig: ttt_rr_chem_acc} that many molecular systems start with large errors and these errors are decreased to well within $40$ meV / Å with TTT and RR. Additionally, hundreds of molecules across a range of errors have errors that are brought down significantly closer to the in-distribution performance. These results suggest that MLFFs have the expressivity to model more diverse chemical spaces, and can be better trained to do so.}

\paragraph{Evaluating Improved Downstream Utility.} We run reference DFT simulations to ensure that the improved stability reported in \tref{tab: spice_ttt_indiviual} and \tref{tab: spice_rr_indiviual} translate into improved simulation quality in terms of the predicted distribution of interatomic distances h(r) (see \sref{exp:sim_md17} and \sref{apx:exp_details} for details). Due to the computational cost of running reference MD simulations, we are only able to do this on a subset of the molecules. RR on top of MACE-OFF lowers the h(r) MAE from $0.17$ and $0.09$ to $0.15$ and $0.07$ for $SC_6H_4$ and $IC_2H$, respectively. TTT on top of GemNet-T lowers the h(r) MAE from $0.20,$ $0.44,$ and $0.39$ to $0.18,$ $0.19,$ and $0.17$ for $N_3C_5H_3$, $IC_2H$, and $O_3P$, respectively. 

We run also BFGS structure relaxations with the MLFFs and caclulate how many extra steps are needed with DFT to find a relaxed structure. GemNet-T requires 10.4 additional steps on average (on $N_3C_5H_3$, $IC_2H$, and $O_3P$, 3 seeds each), whereas GemNet-T+TTT requires 7.7. MACE-OFF requires 4.0, and MACE-OFF+RR requires 3.8 (on $SC_6H_4$ and $IC_2H$, 3 seeds each).

\begin{table}[]
\centering
\setlength{\tabcolsep}{2pt}
\begin{tabular}{llccccccc}
\toprule
 &  & \textbf{Overall} & \textbf{\begin{tabular}[c]{@{}c@{}}$O_2ClSNC_{8}$-\\ $H_{16}$\end{tabular}} & \textbf{\begin{tabular}[c]{@{}c@{}}$O_2N_2C_{16}$-\\ $SH_{14}$\end{tabular}} & \textbf{\begin{tabular}[c]{@{}c@{}}$O_3C_{19}$-\\ $SiH_{26}$\end{tabular}} & \textbf{\begin{tabular}[c]{@{}c@{}}$O_2N_2C_{16}$-\\ $SiH_{28}$\end{tabular}} & \textbf{\begin{tabular}[c]{@{}c@{}}$Cl_2C_7$-\\ $SiH_{14}$\end{tabular}} & \textbf{\begin{tabular}[c]{@{}c@{}}$Cl_3C_9$-\\ $SiH_{11}$\end{tabular}} \\ \hline
\multirow{2}{*}{\textbf{\begin{tabular}[c]{@{}l@{}}Force MAE\\ \\ (meV / Å)\end{tabular}}} & \textbf{GemNet-dT} & $78.3 \pm 7.8$ & 38 & 33 & 74 & 75 & 109 & 107 \\ 
 & \textbf{\begin{tabular}[c]{@{}l@{}}GemNet-dT \\ + TTT\end{tabular}} & $56.6 \pm 5.6$ & 28 & 26 & 35 & 39 & 46 & 44 \\ \bottomrule 
\end{tabular}
\caption{\update{\textbf{Test-Time Training (TTT) with a Semi-Empirical Prior on SPICEv2 Benchmark.} We evaluate a GemNet-T model trained on 951k samples from SPICE on a held-out set of 10k new molecules from SPICEv2. To evaluate the effectiveness of TTT, we use the semi-empirical GFN2-xTB~\citep{semiempirical} as a prior and apply TTT to our SPICEv2 distribution shift benchmark. The results show that TTT with a semi-empirical prior improves performance across a range of error levels, bringing many molecules close to the performance achieved on in-distribution data. We report 95\% confidence intervals for the overall error on the entire test set and highlight individual molecule examples to illustrate the benefits of TTT.
}} \label{tab:semiempirical}
\end{table}

\update{\paragraph{TTT is agnostic to the chosen prior.} We explore using the semi-empirical GFN2-xTB \citep{semiempirical} as the prior to provide further evidence that TTT is agnostic of the prior chosen. We train a GemNet-dT model with the pre-train, freeze, fine-tune approach described in \sref{sec: ttt} using GFN2-xTB as the prior. The results in \tref{tab:semiempirical} show that TTT with GFN2-xTB also enables better performance across a range of errors.}

\begin{figure}[H]
    \centering
    \includegraphics[width=0.9\linewidth]{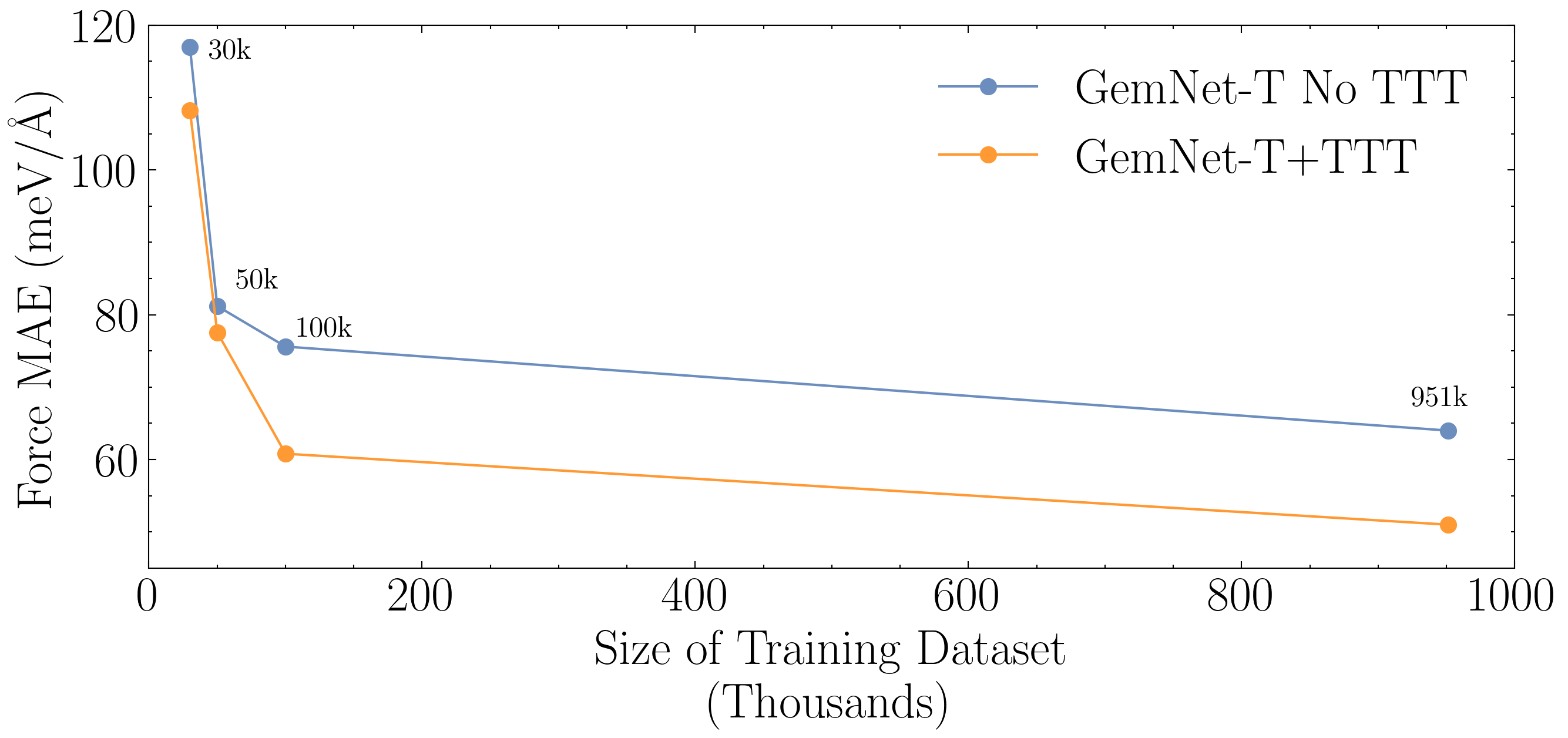}
    \caption{\update{\textbf{Performance on the SPICEv2 Distribution Shift Benchmark Versus Dataset Size.} We evaluate GemNet-T models trained on increasing amounts of data from SPICE on 10k new molecules from SPICEv2. The results show that while increasing the training dataset size improves performance on the SPICEv2 benchmark, the gains in accuracy diminish rapidly. Test-Time Training (TTT) consistently improves performance across all dataset sizes.}}
    \label{fig:err_v_trainingdata}
\end{figure}

\paragraph{Scaling Experiment on SPICEv2: Investigating the Impact of Dataset Size on Out-of-Distribution Performance.} We conduct a scaling experiment to understand out-of-distribution performance with and without TTT as a function of dataset size. We train four GemNet-T models on different subsets of the SPICE dataset: 30k, 50k, 100k, and the full 951k samples. 
Our results, presented in \fref{fig:err_v_trainingdata}, show that increasing the dataset size improves generalization performance on SPICEv2, but with diminishing returns. This suggests that simply adding more in-distribution data may not be sufficient to achieve optimal generalization performance, consistent with our findings in \fref{fig:foundation_bad} and \sref{sec:distribution_shifts}. Notably, TTT consistently improves performance across all dataset sizes, and the benefits of TTT do not decrease even when using the full 951k dataset.

\subsection{Additional Results on MD17}
\label{apx:nve_md17}

\update{We additionally run NVE simulations \citep{fu2023forces, fu2025learningsmoothexpressiveinteratomic} with the Velocity Verlet integrator \citep{HjorthLarsen2017ase} before and after TTT. As with the NVT simulations, we use a $0.5$ fs time step and simulate for $100$ps. Although simulations on naphthalene are slightly more unstable, TTT still increases the stability of simulations (see \tref{tab:stable_nve}).}

\begin{figure}
    \centering
    \includegraphics[width=0.75\linewidth]{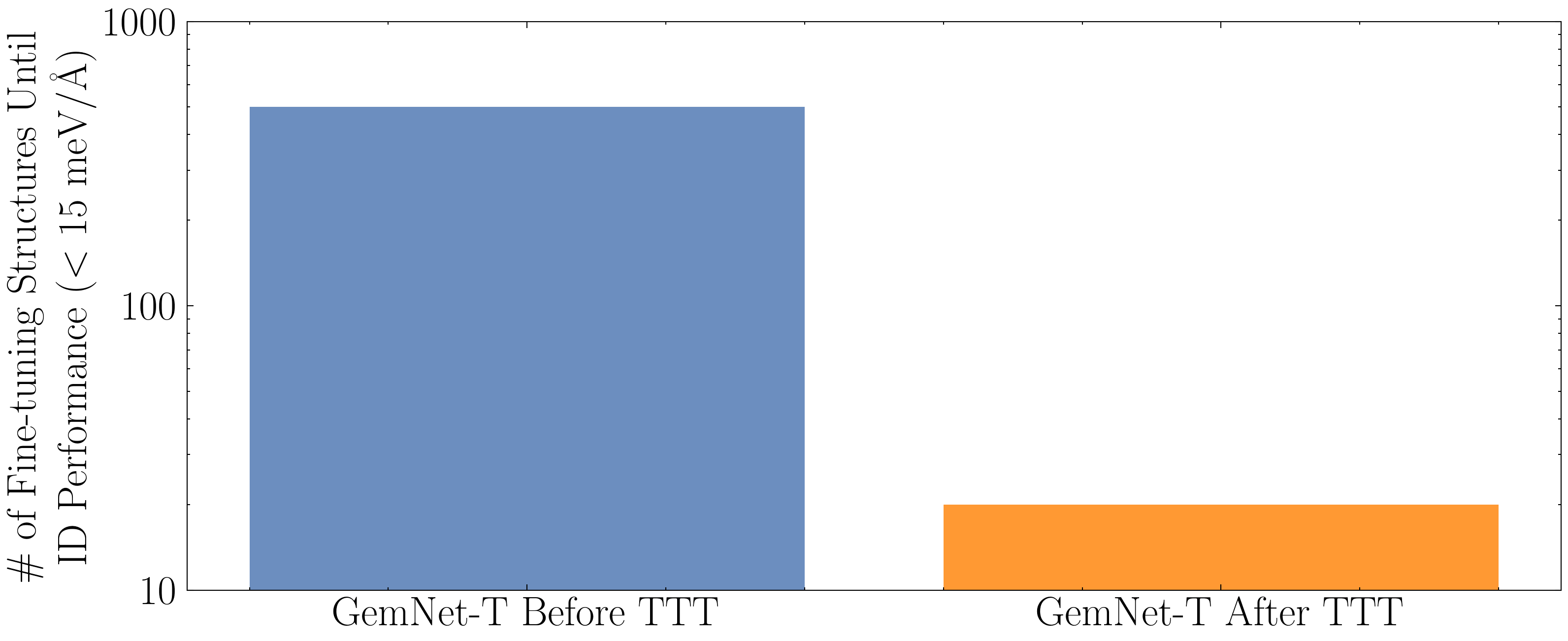}
    \caption{\textbf{Test-Time Training (TTT) Improves Fine-Tuning Efficiency on MD17 dataset.} We demonstrate the effectiveness of TTT in reducing the amount of data required for fine-tuning a GemNet-dT model to achieve in-distribution performance. Initially, we train the model on a small set of three molecules from the MD17 dataset. We then fine-tune the model on a new, unseen molecule (toluene) with and without TTT. Our results show that applying TTT before fine-tuning enables the model to reach in-distribution performance ($<15$ meV / Å) with 10 times less data compared to fine-tuning without TTT.
    }
    \label{fig:ft_after_ttt}
\end{figure}

We also demonstrate that TTT can be used in conjunction with fine-tuning. We fine-tune the GemNet-dT model used in \sref{exp:sim_md17} on the out-of-distribution toluene molecule. We measure how much data is needed to reach the in-distribution performance of less than $15$ meV / Å. This fine-tuning is done both before and after TTT is conducted. \fref{fig:ft_after_ttt} shows that TTT provides a much better starting point for fine-tuning, reducing the number of reference labels needed to reach the in-distribution performance by more than $20\times$.

\begin{table}[]
\centering
\begin{tabular}{lll}
\toprule
\textbf{Molecule} & \textbf{GemNet-T} & \textbf{GemNet-T + TTT} \\ \hline
Toluene & \textless{}1ps & 100 $\pm$ 0 ps\\ \hline
Naphthalene & \textless{}1ps & 43 $\pm$ 5.2 ps \\ \bottomrule
\end{tabular}
\caption{\update{\textbf{Stability of NVE Simulations with Test-Time Training (TTT).} We train a GemNet-dT model on three molecules from MD17 and evaluate its ability to simulate new molecules not seen during training. TTT enables stable NVE simulations for molecules unseen during training. We report mean $\pm$ standard deviation across 3 seeds.}}\label{tab:stable_nve}
\end{table}

\subsection{Test-Time Radius Refinement with JMP on ANI-1x}
\label{apx:ttrr_jmp}

\update{We evaluate whether our proposed test-time radius refinement (RR) method (see \ref{sec: ttrr}) can help JMP \citep{shoghi2023jmp} address connectivity distribution shifts in the ANI-1x dataset \citep{Smith2020_ani1x}. Following the approach outlined in \sref{exp:spice_maceoff}, we search over $7$ different radius cutoffs from $6.5$ to $9.5$ Å to find the one that best matches the training Laplacian eigenvalue distribution.}

\update{As shown in \tref{tab:ttrr_jmp} and \tref{tab:ttrr_jmp_individual}, RR is able to improve force errors for JMP, including improving errors that are already low. We again highlight the top 10\% of molecules with the greatest improvement, since the improvements from RR are right-skewed. RR often improves errors by 10-20\% for individual molecules. This experiment provides further evidence that RR can address connectivity distribution shifts for existing pre-trained models at minimal computational cost, suggesting that existing models overfit to the graph structures seen during training.}

\begin{table}[]
\centering
\begin{tabular}{lllllll}
\toprule
 & \multicolumn{3}{l}{\textbf{Force Error Range (meV / Å)}} & & & \\
 & \multicolumn{2}{c}{0-43}& \multicolumn{2}{c}{43-100}& \multicolumn{2}{c}{\textgreater{}100}\\ \hline
\begin{tabular}[c]{@{}l@{}}\textbf{JMP on ANI-1x Test Set (Top 10\%)} \\ Force MAE (meV/Å)\end{tabular} 
& \multicolumn{2}{c}{\begin{tabular}[c]{@{}c@{}}$17.4 \pm 0.02$ \\ $(15.1 \pm 0.07)$\end{tabular}} 
& \multicolumn{2}{c}{\begin{tabular}[c]{@{}c@{}}$52.4 \pm 0.18$ \\ $(52.3 \pm 0.54)$\end{tabular}} 
& \multicolumn{2}{c}{\begin{tabular}[c]{@{}c@{}}$151.7 \pm 8.4$ \\ $(167.7 \pm 39.3)$\end{tabular}} \\

\begin{tabular}[c]{@{}l@{}} \textbf{JMP + RR (ours) on ANI-1x Test Set (Top 10\%)} \\ Force MAE (meV/Å)\end{tabular} 
& \multicolumn{2}{c}{\begin{tabular}[c]{@{}c@{}}$17.3 \pm 0.02$ \\ $(14.6 \pm 0.07)$\end{tabular}} 
& \multicolumn{2}{c}{\begin{tabular}[c]{@{}c@{}}$52.3 \pm 0.18$ \\ $(51.9 \pm 0.54)$\end{tabular}} 
& \multicolumn{2}{c}{\begin{tabular}[c]{@{}c@{}}$151.5 \pm 8.3$ \\ $(163.6 \pm 37.8)$\end{tabular}} \\
\bottomrule
\end{tabular}
\caption{\update{\textbf{Test-Time Radius Refinement with JMP on ANI-1x.} We implement our test-time radius refinement method (see \sref{sec: ttrr}) on JMP and evaluate improvements on the ANI-1x test set defined in \citet{shoghi2023jmp}. Test-time radius refinement helps improve performance by mitigating connectivity distribution shifts. We highlight the top 10\% of molecules with the greatest improvement in parentheses to show that test-time radius refinement helps across a range of errors.}} \label{tab:ttrr_jmp}
\end{table}

\begin{table}[]
\centering
\begin{tabular}{lllllll}
\toprule
\multicolumn{7}{c}{\textbf{Example Molecules Force MAE Before $\rightarrow$ After RR} (meV / Å)}\\
$C_3H_{10}N_2O_2$ & $C_5H_3NO $ & $C_5H_6N_2O$ & $C_5H_5NO_2$ & $C_5H_3N_3$ & $C_3H_6O_2$ & \\
$6.9 \rightarrow 5.4$ & $8.2 \rightarrow 6.2$ & $53.0 \rightarrow 44.2$ & $85.2 \rightarrow 78.3$ & $101.1 \rightarrow 99.7$ & $158.9 \rightarrow149.7$ & \\ \bottomrule
\end{tabular}
\caption{\update{\textbf{Individual Examples from ANI-1x with Radius Refinement (RR) on JMP.} We perform RR when evaluating JMP on molecules from the ANI-1x test set. We highlight individual molecular examples to show that RR helps across a range of errors.}} \label{tab:ttrr_jmp_individual}
\end{table}

\subsection{Evaluating Distribution Shifts in the MD22 Dataset: Low to High Force Norms}

\label{exp:h2l_md22}
We establish a benchmark for force norm distribution shifts, using the MD22 dataset \citep{chmiela_accurate_2023}. The MD22 data set contains large organic molecules with samples generated by running constant-temperature (NVT) simulations, meaning that the majority of the structures are in lower energy states, and thus have low force norms. We filter out structures that have an average per-atom force norm smaller than a $1.7$ eV / Å cutoff, which filters out about half of the data. We then evaluate whether GemNet-dT can generalize to high-force norm structures. 

We train three different GemNet-dT models on 3 MD22 molecules—Ac-Ala3-NHMe, stachyose, and buckyball catcher—using the filtered low force norm dataset. We evaluate the GemNet-dT model on structures with force norms larger than the training cutoff. We also perform TTT using sGDML as the prior, as described in \sref{sec: ttt}, to mitigate the distribution shift on the high-force norm test samples. For more details, see \sref{apx:exp_details}.

\paragraph{Force Norm Generalization Results.} As shown in \tref{tab:md22_l2h}, GemNet-dT performs poorly on high force norm structures when compared to the low force norm structures it sees during training. TTT can mitigate the force norm distribution shift and close the gap between the in-distribution and out-of-distribution performance. This result further supports the hypothesis that MLFFs struggle to learn generalizable representations even when facing a distribution shift in a narrow single molecule dataset.

\begin{table}[]
\centering
\setlength{\tabcolsep}{3pt}
\begin{tabular}{ccccc}
\toprule
\textbf{Force Norm} & \textbf{} & \multicolumn{3}{c}{\textbf{Force MAE (meV / Å)}}\\
\textbf{Average} & \textbf{Model} & \multicolumn{1}{l}{Ac-Ala3-NHMe} & \multicolumn{1}{l}{Stachyose} & \multicolumn{1}{l}{Buckyball Catcher}  \\ \hline
\textless 1.7 eV / Å & GemNet-dT & 11.6 & 11.7 & 8.7 \\ \hline
\textgreater 1.7 eV / Å & \begin{tabular}[c]{@{}c@{}}GemNet-dT\\ $\downarrow$\\ GemNet-dT + TTT\end{tabular} & \begin{tabular}[c]{@{}c@{}}36.8\\ $\downarrow$\\ 26.5\end{tabular} & \begin{tabular}[c]{@{}c@{}}24.2\\ $\downarrow$\\ 19.0\end{tabular} & \begin{tabular}[c]{@{}c@{}}16.4\\ $\downarrow$\\ 12.7\end{tabular} \\ \bottomrule
\end{tabular}
\caption{\textbf{Evaluating Low to High Force Norms on MD22.} We train a GemNet-dT model on low force norm structures from MD22 ($<1.7$ eV / Å force norm averaged over atoms) and evaluate the model on high force norm structures ($>1.7$ eV / Å). GemNet-dT generalizes poorly to the high force norm structures, but TTT significantly closes the gap.}
\label{tab:md22_l2h}
\end{table}

\subsection{Test-Time Training on OC20}
\label{sec:oc20ttt}

The Open Catalyst 2020 (OC20) dataset consists of relaxation trajectories between adsorbates and surfaces~\citep{Chanussot_2021_oc20}. The primary training objective consists of mapping structures to their corresponding binding energy and forces (S2EF), as determined by DFT calculations. Both the S2EF task and OC20 dataset are challenging, due to the diversity in atom types and system sizes. The OC20 dataset includes an out-of-distribution test split consisting of systems that were not encountered during training. Even models trained on the full 100M+ OC20 dataset perform significantly worse on the out-of-distribution split \citep{Chanussot_2021_oc20}. Consistent with previous test-time training work \citep{sun2020testtime, gandelsman2022testtime, jang2023testtime}, we use this split to assess our TTT approach.

\paragraph{Problem Setup.}
For our prior, we use the Effective Medium Theory (EMT) potential, introduced by \citet{JACOBSEN_emt}. Using this, we can compute energies and forces for thousands of structures in under a second using only CPUs~\citep{HjorthLarsen2017ase}. The EMT potential currently only supports seven metals (Al, Cu, Ag, Au, Ni, Pd and Pt), as well as very weakly tuned parameters for H, C, N, and O. Consequently, we filter the 20 million split in the OC20 training dataset to only the systems with valid elements for EMT, leaving 600 thousand training examples. Similarly, the validation split is filtered and reduced to 21 thousand examples. While this work primarily focuses on evaluating our TTT approach, exploring the potential of a more general prior, or developing such a prior, represents a promising direction for future work.

\paragraph{Training Procedure.}

We use a joint training loss function, $\mathcal{L} = \mathcal{L}_P + \mathcal{L}_M$,
to train a GemNet-OC model \citep{gasteiger2022gemnetoc}, which is specifically optimized for the OC20 dataset. At test-time, we use the EMT potential to label all structures with forces and total energies. For each relaxation trajectory in the validation dataset, we update our representation parameters with the prior objective, $\mathcal{L}_P$ (see~\eref{eqn:ttt}), and then make predictions with the updated parameters (see~\eref{eqn:ttt_pred}). The TTT updates are performed individually for each system in the validation set. See \tref{tab:ttthyperoc20} for hyperparameters.

\paragraph{Results.}

\begin{figure}
    \centering
    \includegraphics[height=200pt]{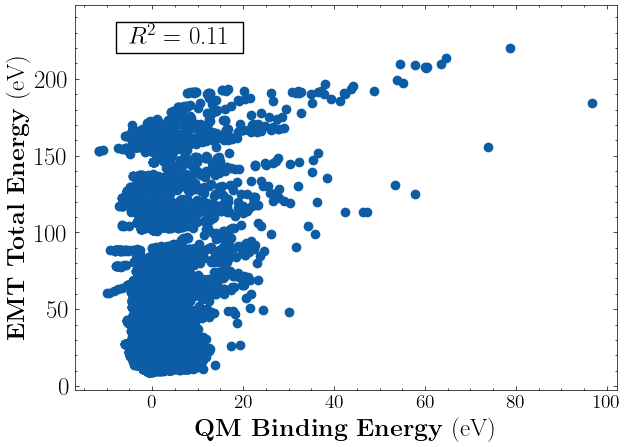}
    \caption{\textbf{EMT Correlation with Reference Energy DFT Calculations on OC20.} We compare the DFT energy to the predicted energy from the EMT prior on samples from OC20. The correlation is very weak.}
    \label{fig:emt_acc}
\end{figure}

We compare the performance of our joint-training plus TTT method against a baseline GemNet-OC model trained only on DFT targets and evaluated without TTT on the validation set. Despite the weak correlation between EMT labels and the more accurate DFT labels (see  \fref{fig:emt_acc}), using EMT labels for joint-training helps regularize the model and improves performance on the out-of-distribution split. After joint-training, implementing test-time training steps further improves the model's performance (see \tref{table:oc20_ttt}). This demonstrates that even though EMT has limited predictive accuracy as a prior, it can still be used to learn more effective \textit{representations} that generalize to out-of-distribution examples. This experiment provides further evidence that improved training strategies can help existing models address distribution shifts.    

\begin{table}
  \caption{\textbf{OC20 test-time Training.} We evaluate a GemNet-OC model on the OC20 out-of-distribution validation split to assess the impact of joint-training and TTT. The model is trained on 600 thousand examples from the OC20 20M split that have elements supported by the EMT prior.}
  \label{table:oc20_ttt}
  \centering
  \begin{tabular}{lll}
    \toprule
    \cmidrule(r){2-3}
    \textbf{Model}     & \textbf{Force MAE} (meV/\AA)& \textbf{Energy MAE} (meV)\\
    \midrule
    GemNet-OC & 77.8& 1787.4\\
    GemNet-OC Joint Training (ours)& 63.67& 1320\\
    GemNet-OC Joint Training + TTT (ours) & \textbf{61.42}& \textbf{1143}\\
    \bottomrule
  \end{tabular}
\end{table}

\begin{table}
\caption{\textbf{TTT Hyperparameters for OC20 OOD Split.}}\label{tab:ttthyperoc20}
\centering
\begin{tabular}{ll}
\toprule
\textbf{Hyperparameter} & \textbf{Value} \\ \hline
Steps                   & 11           \\
Learning Rate           & 1e-4           \\
Optimizer               & Adam                       \\
Weight Decay            & 0.001          \\ \bottomrule
\end{tabular}
\end{table}

\subsection{Additional Potential Energy Surfaces Before and After Test-Time Training}

\begin{figure}
\centering
\begin{subfigure}{0.25\textwidth}
  \centering
\includegraphics[width=\linewidth]{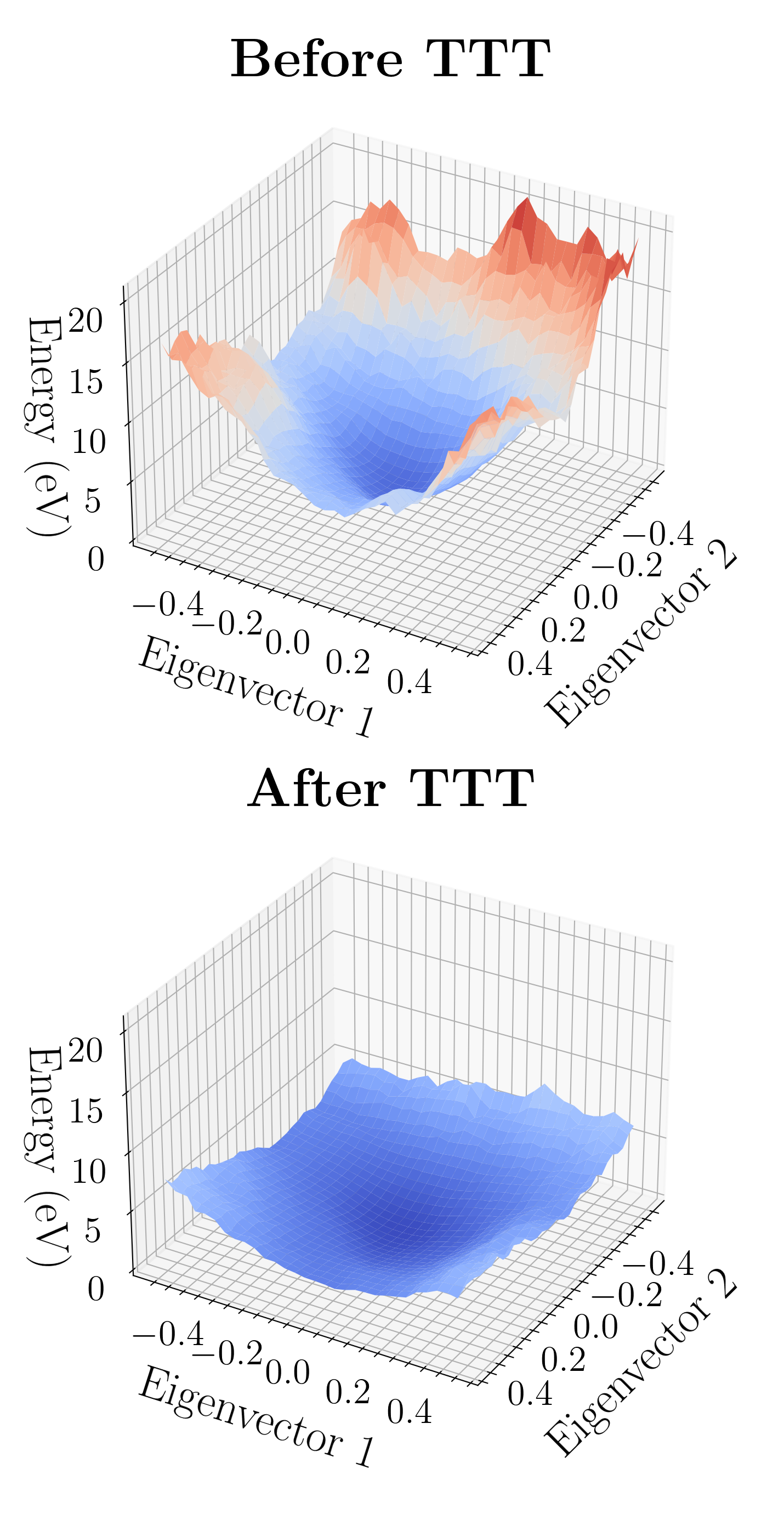}
  \caption{Toluene}
\end{subfigure} 
\begin{subfigure}{0.25\textwidth}
  \centering
\includegraphics[width=\linewidth]{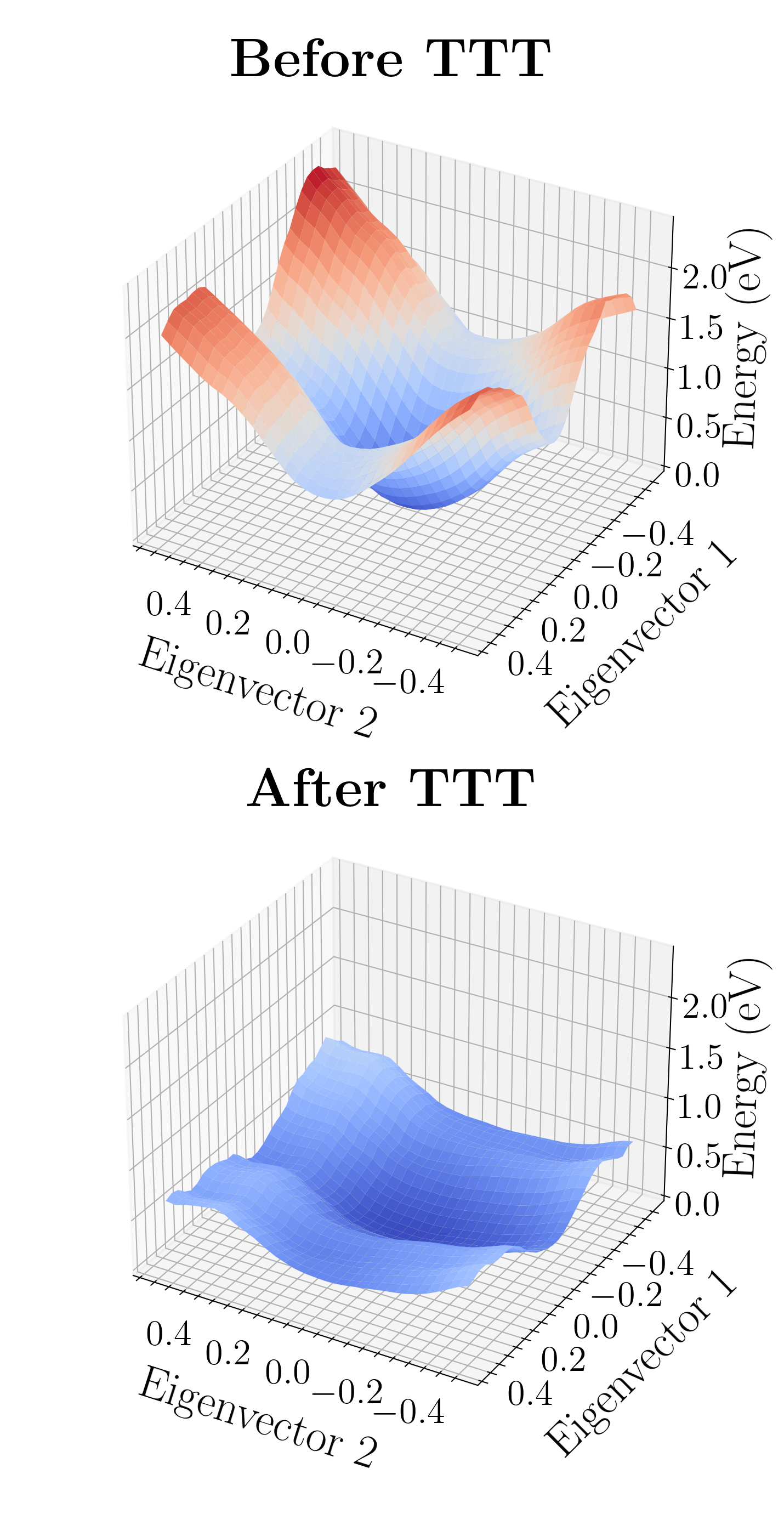}
  \caption{AT-AT}
\end{subfigure}
\begin{subfigure}{0.25\textwidth}
    \centering
\includegraphics[width=\linewidth]{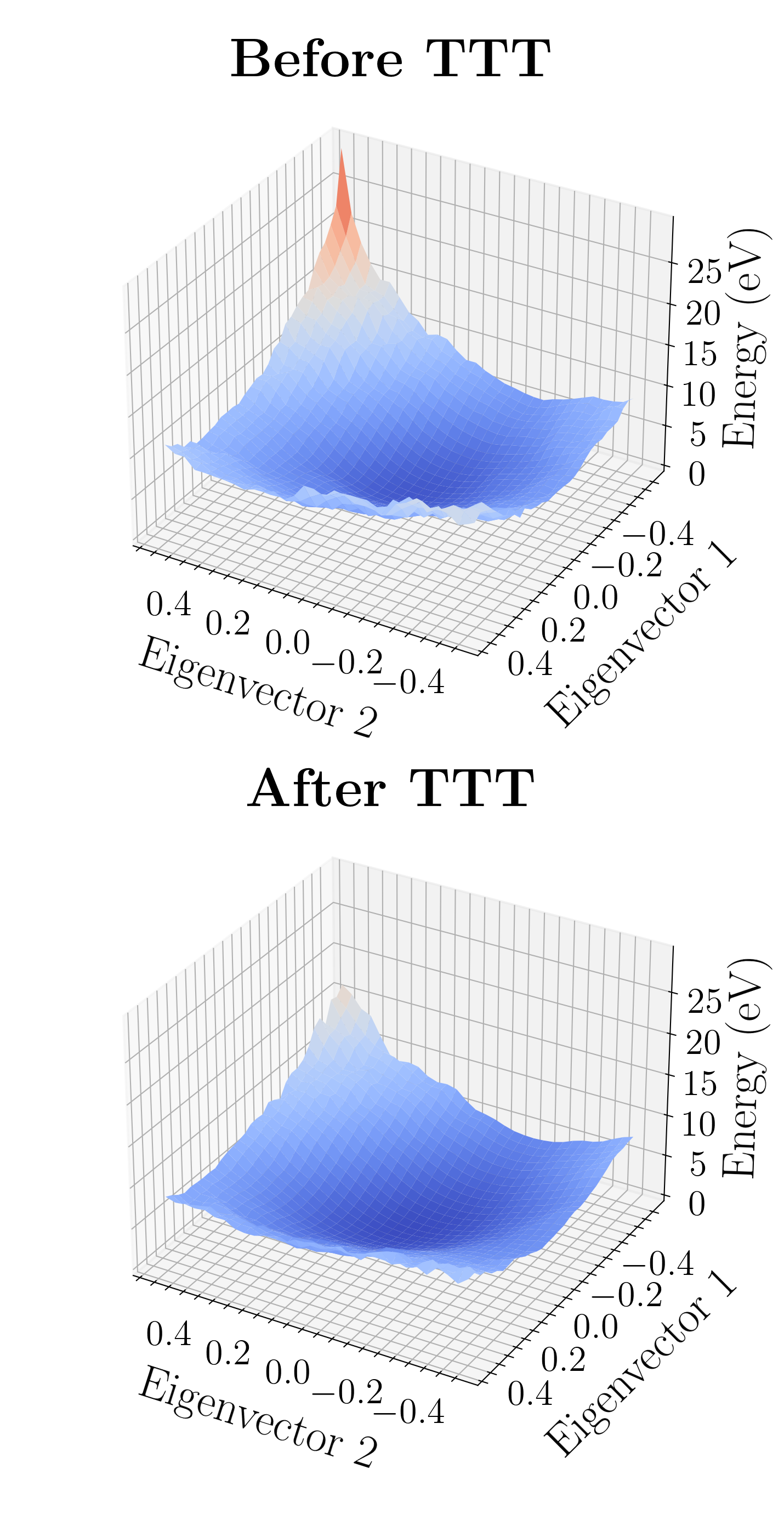}
    \caption{Naphthalene}
\end{subfigure}
\caption{\textbf{Predicted Potential Energy Surfaces for Molecules in MD17 and MD22.} We consider a GemNet-dT model trained on three molecules from MD17. We plot the predicted potential energy surface, before and after test-time-training, from the model along the first two principal components of the Hessian for new molecules not seen during training. TTT regularizes the model and smooths the predicted potential energy surface.}
\label{fig: more_pes}
\end{figure}

We provide additional potential energy surface plots in \fref{fig: more_pes}. TTT consistently smooths the predicted potential energy surface. We plot the energy along the two principal components of the energy Hessian.

\section{Experiment Details}
\label{apx:exp_details}

We describe in detail the  benchmarks established in this paper along with experiment hyperparameters. Code for benchmarks and training methods will be made available.

In line with previous test-time training works \citep{sun2020testtime, gandelsman2022testtime, jang2023testtime}, we update as few parameters as possible during TTT. For MD17, MD22, and SPICE experiments, we train everything before the second interaction layer in GemNet-T/dT. For OC20 (see \sref{sec:oc20ttt}), we train everything before the second output block in GemNet-OC. 

Hyperparameters were largely adapted from \citet{fu2023forces}, although we increased the batch size to 32 to speed up training for GemNet-dT. Other deviations from \citet{fu2023forces} are mentioned below. 

\subsection{SPICEv2 Distribution Shift Benchmark}

\paragraph{Dataset Details.} We evaluate models trained on MACE-OFF's training split \citep{kovács2023maceoff23}, consisting of 951k structures primarily from the SPICE dataset \citep{Eastman2023spice}. The test set contains 10,000 new molecules from SPICEv2 \citep{eastman2024spice2} not seen in the MACE-OFF training split. The 10,000 molecules were chosen to be the molecules that had the most structures in order to provide a large test set of 475,761 structures. GemNet-T was trained on the same data as MACE-OFF.


\vspace{-10pt}
\update{\paragraph{Simulation Details.}
We run simulations for 100 ps using a temperature of 500K and a Langevin thermostat (with friction 0.01), otherwise following the parameters used in \citet{fu2023forces}. Since the SPICEv2 dataset was not generated purely from MD simulations, we do not have reference $h(r)$ curves for this dataset and instead focus on stability.}

\paragraph{Hyperparameters.} Hyperparameters were adapted from \citet{fu2023forces}, with the following modifications shown to scale the model to 4M parameters to be more in line with MACE-OFF's 4.7M parameters:
\begin{enumerate}
    \item Atom Embedding Size: $128 \rightarrow 256$
    \item RBF Embedding Size: $16 \rightarrow 32$ 
    \item Epochs: 250
\end{enumerate} 
For test-time training parameters, see \tref{tab:ttt_params_spice}. Note that we performed early stopping if the prior loss got stuck, or if it reached the in-distribution loss (since this implies overfitting and deteriorates performance on the main task).

\begin{table}[]
\centering
\begin{tabular}{ll}
\toprule
\textbf{Parameter} & \textbf{Value} \\ \hline
Learning Rate & 1e-4 \\
Momentum & 0.9 \\
Optimizer & SGD \\
Weight Decay & 0.001 \\
Steps & 250 \\
\bottomrule
\end{tabular}
\caption{\textbf{TTT Parameters for SPICEv2 Distribution Shift Benchmark.}} \label{tab:ttt_params_spice}
\end{table}

\subsection{Assessing Low to High Force Norms on MD22}

\paragraph{Dataset Details.} We train on approximately 6k samples from each molecule, corresponding to the 10\% split for Ac-Ala3-NHME, 25\% for stachyose, and 100\% for buckyball catcher. 

\begin{figure}
    \centering
    \includegraphics[width=\linewidth]{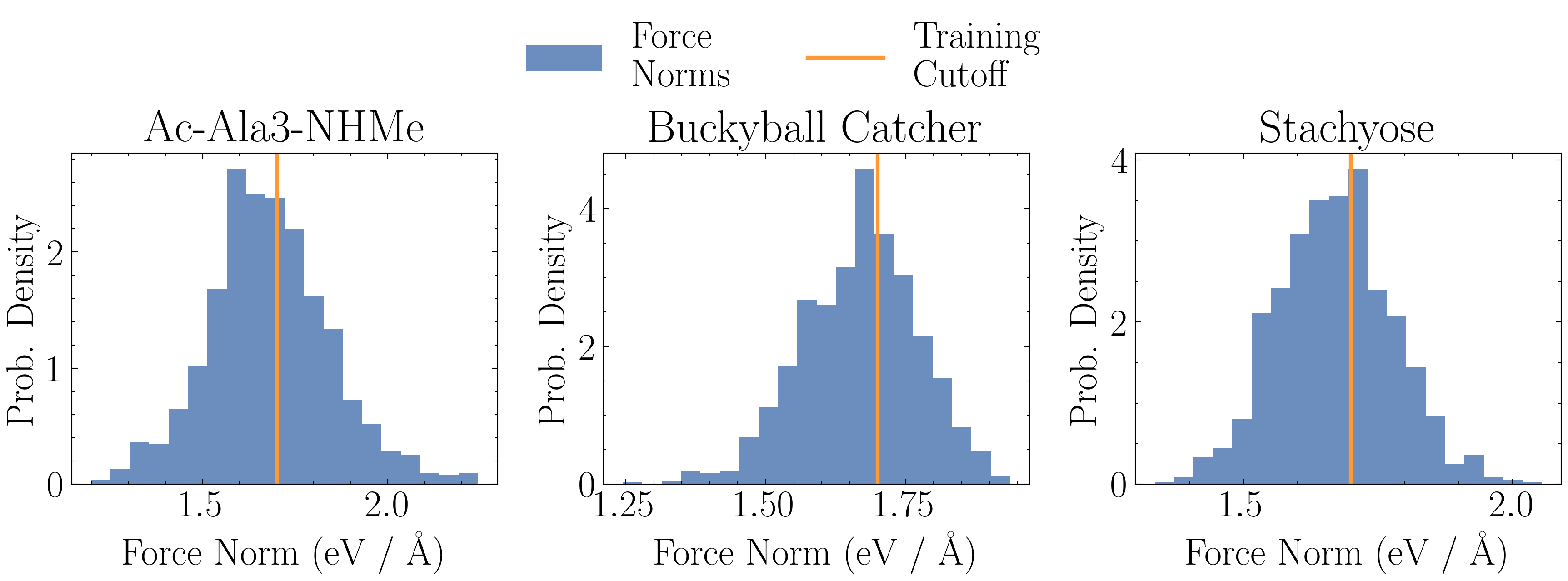}
    \caption{\textbf{Force Norms for MD22 Force Norm Distribution Shift Experiment.} We plot the force norms for molecules from the MD22 dataset. The line in orange indicates the force norm cutoff used to train the models in \sref{exp:h2l_md22}. Note that since the dataset was generated with NVT simulations, force norms are generally low when compared to SPICE.}
    \label{fig:enter-label}
\end{figure}

\paragraph{Hyperparameters.} See \tref{tab:md22ttthyper} for details on the hyperparameters used.
\begin{table}

\caption{\textbf{TTT Hyperparameters MD22 Experiments.} We note that especially in cases where the prior is reasonably accurate, TTT is generally robust to a wide range of hyperparameter choices.}\label{tab:md22ttthyper}
\centering
\begin{tabular}{ll}
\toprule
\textbf{Hyperparameter} & \textbf{Value} \\ \hline
Steps                   & 50 \\
Learning Rate           & 1e-5           \\
Optimizer               & SGD            \\
Momentum                & 0.9            \\
Weight Decay            & 0.001          \\ \bottomrule
\end{tabular}
\end{table}

\subsection{Simulating Unseen Molecules on MD17}

We provide further experimental details for the simulating unseen molecules benchmark on MD17 (see \sref{exp:sim_md17}).

\paragraph{Dataset Details.} We use the 10k dataset split for the 3 training molecules (aspirin, benzene, and uracil). For test-time training, the 1k test-set is used for naphthalene and toluene. We note that TTT can also be done with  structures generated from simulations with the prior, and we think further experimentation with this is an interesting direction for future work. 

\paragraph{Simulation Details.}
We run simulations for 100 ps using a temperature of 500K and a Langevin thermostat (with friction 0.01), otherwise following the parameters used in \citet{fu2023forces}. We measure the distribution of interatomic distances $h(r)$ to evaluate the quality of the simulations. The distribution of interatomic distances is defined as:
\begin{equation}
    \label{eqn:hrdef}
    h(r) = \frac{1}{n (n-1)} \sum_{i}^n \sum_{j \neq i}^n \delta (r - ||\mathbf{x_i} - \mathbf{x_j}||),
\end{equation}
where $r$ is a reference distance, $\mathbf{x_i}$ denotes the position of atom $i$, $n$ is the total number of atoms, and $\delta$ is the Dirac Delta function. The MAE between a predicted $\hat{h}(r)$ and a reference $h(r)$ is given by:
\begin{equation}
    \label{eqn:hrmae}
    \text{MAE}(\hat{h}(r), h(r)) = \int_0^\infty |\langle h(r) \rangle - \langle \hat{h}(r) \rangle| dr,
\end{equation}
where $\langle \cdot \rangle$ indicates time averaging over the course of the simulation.

In both cases, TTT brings down force errors from $\sim200$ meV / Å down to less than $25$ meV / Å, beating the prior (that uses 50 samples) and enabling stable simulation. We found that a prior that uses only 15 samples still leads to improvements with TTT (see \fref{fig:prior_vs_ttt}).

\paragraph{Hyperparameters.} See \tref{tab:ttt_params_md17} for hyperparameters used in the MD17 simulation experiments.

\begin{table}[]
\centering
\begin{tabular}{ll}
\toprule
\textbf{Parameter} & \textbf{Value} \\ \hline
Learning Rate & 1e-3 \\
Momentum & 0.9 \\
Optimizer & SGD \\
Weight Decay & 0.001 \\
Steps & 3000 \\ \bottomrule
\end{tabular}
\caption{\textbf{TTT Parameters for MD17 Transferability Benchmark.}} \label{tab:ttt_params_md17}
\end{table}

\section{Details on Distribution Shifts}
\label{apx:distribution_shifts}

\begin{figure}[H]
    \centering
    \includegraphics[width=0.75\linewidth]{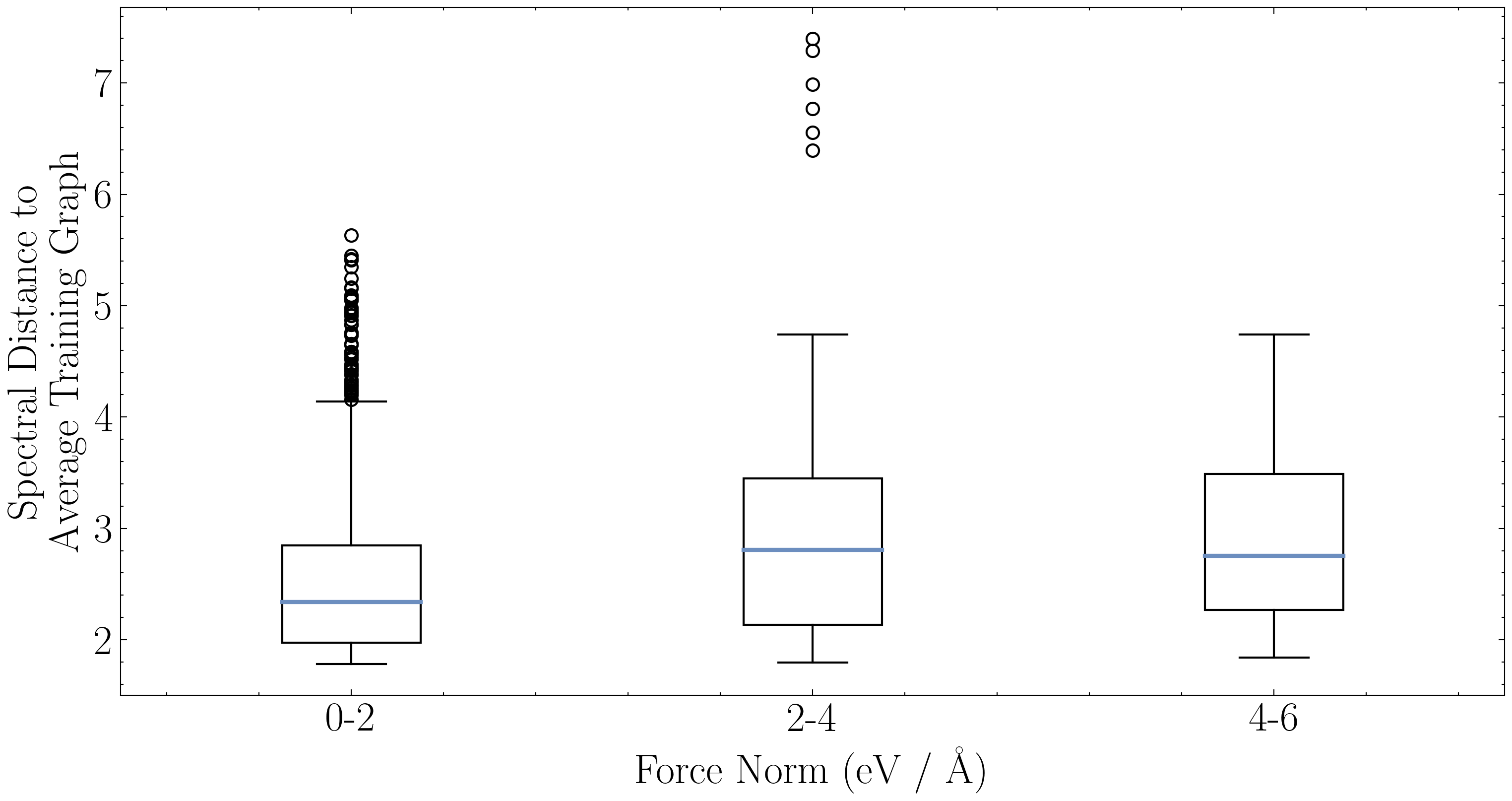}
    \caption{\textbf{Force Norm vs. Connectivity on SPICEv2.} We analyze the force norms and connectivities of new molecules from the SPICEv2 dataset. The distribution of connectivities is similar across force different force norms. This implies that these distribution shifts can happen independently.}
    \label{fig:fn_vs_connect}
\end{figure}

We emphasize that atomic feature, force norm, and connectivity distribution shifts define ``orthogonal" directions along which a shift can happen in the sense that they can each happen independently. In other words, a structure might have the same connectivity and similar force norms, but have a different composition of elements. Similarly,  for the SPICEv2 dataset, the distribution of connectivities is the same independent of force norm of the structure (see \fref{fig:fn_vs_connect}). This implies that one can observe a force norm shift while still seeing similar elements and connectivity.

\begin{figure}[H]
    \centering
    \includegraphics[width=0.3\linewidth]{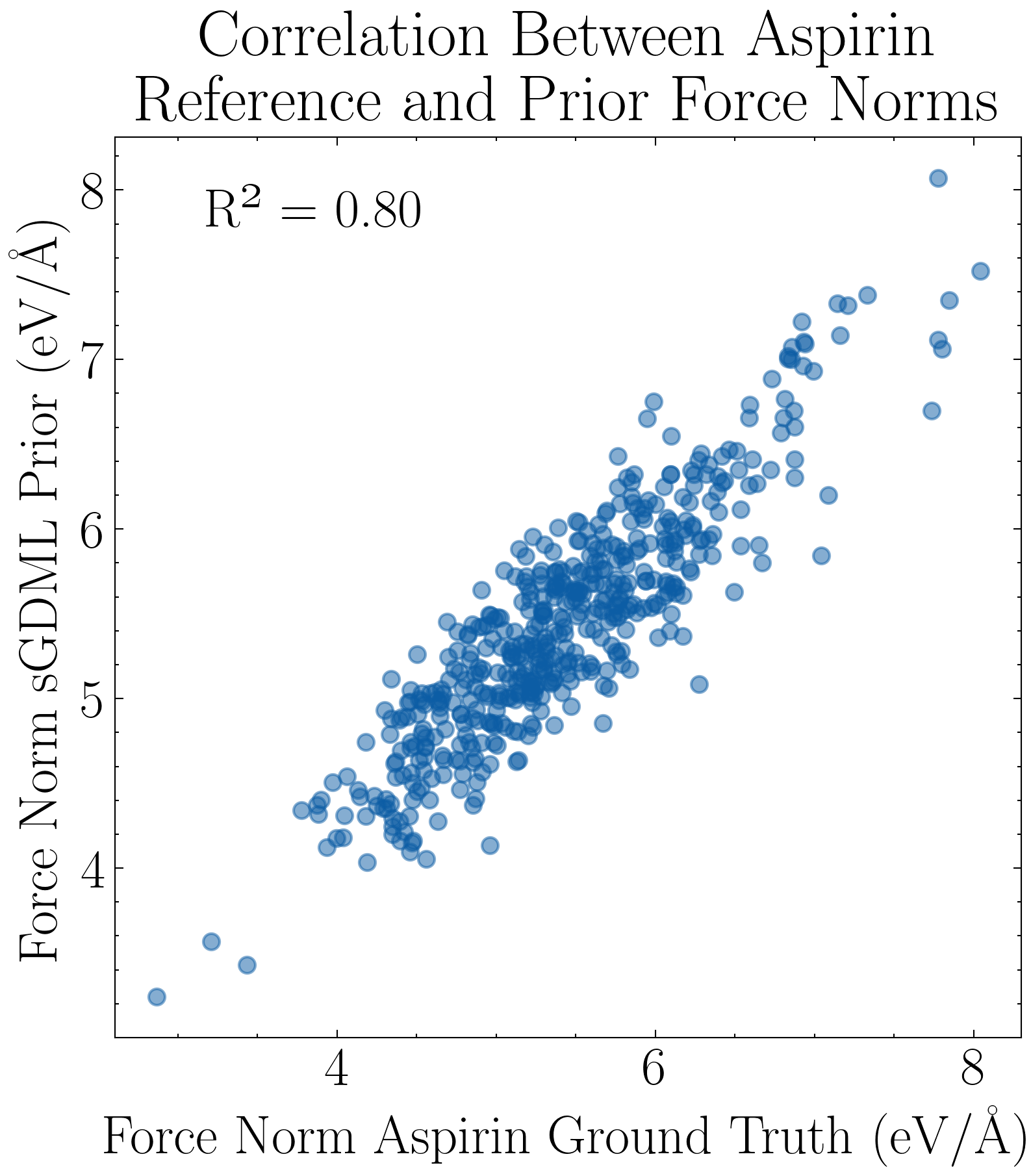}
    \caption{\update{\textbf{Prior and Reference Force Norms are Highly Correlated.} We plot force norms calculated by the sGDML prior and the reference DFT for samples of aspirin from the MD17 dataset. The force norm predicted by the prior is highly correlated with the reference force norm, despite the absolute error between them being large.}}
    \label{fig:prior_force_norm_correlation}
\end{figure}

\begin{figure}[H]
    \centering
    \includegraphics[width=0.65\linewidth]{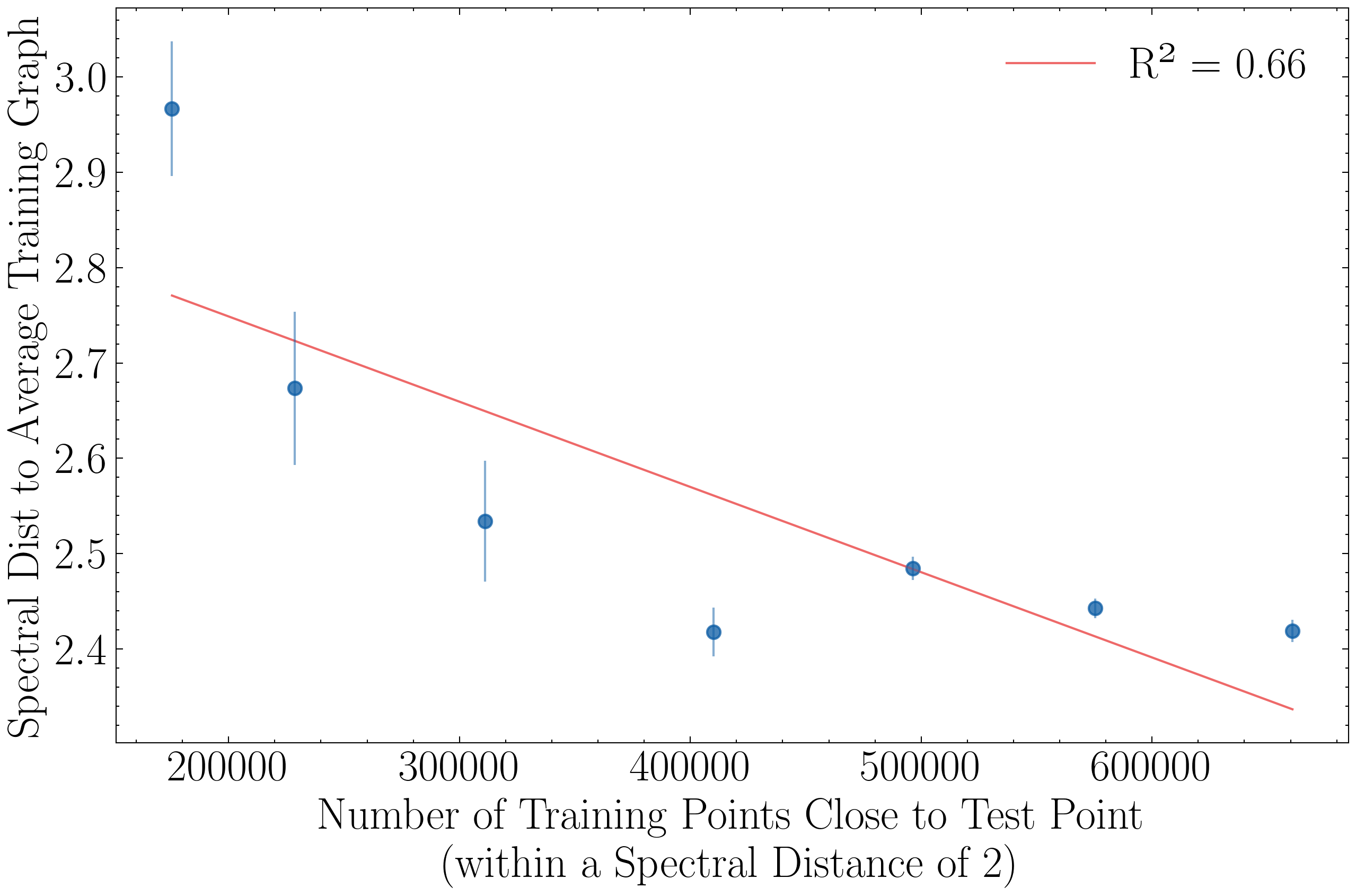}
    \caption{\textbf{Spectral Distance to Average Training Graph Correlates with Number of Training Samples Close to Test Example.} We compare the connectivity of new samples from the SPICEv2 dataset to those seen during training on the SPICE dataset. Although representing the training connectivities with an average Laplacian spectrum is lossy, comparing a test graph to this average spectrum correlates strongly with counting the number of training graphs close to the test graph. $95\%$ confidence intervals are shown with error bars.}
    \label{fig:sn_vs_count}
\end{figure}

\update{Additionally, we provide more details on how we diagnose distribution shifts for new molecules at test time.
\begin{enumerate}
    \item Identifying distribution shifts in the atomic features $\mathbf{z}$ is straightforward: one can simply compare the chemical formula of a new structure to the elements seen during training.
    \item To diagnose force norm distribution shifts, we observe that although priors often have large absolute errors compared to reference calculations, \textit{force norms} are actually highly correlated between priors and reference values (see \fref{fig:prior_force_norm_correlation} for an example from MD17). To determine whether a structure might be out-of-distribution with respect to force norms, the prior can be quickly evaluated at test time, and the predicted force norm can be compared to the training distribution.
    \item Connectivity distribution shifts can be quickly identified by comparing graph Laplacian eigenvalue distributions with the spectral distance (see \ref{sec: ttrr}). Although comparing to the average Laplacian spectra is a lossy representation of the training distribution, comparing individually to all the training graphs is prohibitively expensive in practice. We also observe that counting the number of training graphs close to a test point correlates strongly with the spectral distance between the test graph and the average spectrum (see \fref{fig:sn_vs_count}).   
\end{enumerate}}
\update{We emphasize that our proposed methods for diagnosing distribution shifts are computationally efficient, and they do not require access to reference labels.}

\section{Computational Usage}
\label{apx:compute}
All of our experiments were run on a single A6000 GPU.

\begin{itemize}
    \item 
MD17/22: Training for 100 epochs on a single molecule takes 2 GPU hours. Option 2 from \fref{fig:ttt_compressed} (pre-training, freezing, then fine-tuning) took 2 hours for pre-training and then 2 hours for fine-tuning (although we observed strong finetuning results with even less pre-training). TTT took less than 15 minutes for each molecule. 
\item 
SPICE Results: Pre-training on the prior took less than 5 hours on an A6000 across model sizes. Fine-tuning took 2 days. TTT took less than 5 minutes per molecule. In comparison, MACE-OFF small, medium, and large trained for 6, 10, and 14 A100 GPU-days respectively. Radius refinement takes less than 1 minute per molecule (to calculate eigenvalues to find the optimal radius).
\item 
OC20: Joint-training (option 1) took 48 hours. Evaluation with TTT took 6 hours (compared to 2 hours without TTT).
\end{itemize}


\end{document}